\documentclass{ieeeaccess}
\usepackage{cite}
\usepackage{amsmath,amssymb,amsfonts}
\usepackage{algorithmic}
\usepackage{graphicx}
\usepackage{comment}
\usepackage{textcomp}
\usepackage{algorithm}
\usepackage{algorithmic}
\usepackage{amsmath,amssymb,amsfonts}
\usepackage{algorithmic}
\usepackage{graphicx}
\usepackage{textcomp}
\usepackage{float}
\usepackage{soul}
\usepackage{changepage}
\usepackage{xcolor}
\usepackage{graphicx}
\usepackage{subcaption}
\usepackage{multirow}
\usepackage{xcolor}
\usepackage{comment}
\def\BibTeX{{\rm B\kern-.05em{\sc i\kern-.025em b}\kern-.08em
    T\kern-.1667em\lower.7ex\hbox{E}\kern-.125emX}}

\usepackage{xcolor}
\usepackage{pict2e}
\usepackage{color}
\usepackage{multirow}

\usepackage{enumitem}
\newlist{steps}{enumerate}{1}
\setlist[steps, 1]{label = \textbf {Step \arabic*}:, leftmargin=1.40cm}

\newsavebox{\ORCIDlogo}
\savebox{\ORCIDlogo}{%
\setlength{\unitlength}{\dimexpr 1em/256\relax}%
\begin{picture}(256,256)%
  \color[HTML]{A6CE39}\put(128,128){\circle*{256}}%
  \color{white}%
  \put(78.6,199.2){\circle*{20}}%
  \moveto(70.9,176,9)\lineto(86.3,176,9)\lineto(86.3,69.8)\lineto(70.9,69.8)%
  \closepath\fillpath%
  \moveto(108.9,176.9)\lineto(150.5,176.9)%
  \curveto(190.1,176.9)(207.5,148.6)(207.5 ,123.3)%
  \curveto(207.5,95,8)(186,69.7)(150.7,69.7)%
  \lineto(108.9,69.7)%
  \closepath\fillpath%
  \color[HTML]{A6CE39}%
  \moveto(124.3,83.6)\lineto(148.8,83.6)%
  \curveto(183.7,83.6)(191.7,110.1)(191.7,123.3)%
  \curveto(191.7,144.8)(178,163)(148,163)%
  \lineto(124.3,163)%
  \closepath\fillpath%
\end{picture}%
}

\newcommand\orcidicon[1]{\href{https://orcid.org/#1}{\usebox{\ORCIDlogo}}}

\usepackage{hyperref} 
\hypersetup{
  colorlinks=false,
  linkbordercolor=white,
 urlbordercolor=white,
pdfborder={0 0 0}
} 
\usepackage[switch,columnwise]{lineno}
\begin{document}
\history{Date of publication xxxx 00, 0000, date of current version xxxx 00, 0000.}
\doi{10.1109/ACCESS.2022.0092316}
\title{IoT-Based Real-Time Medical-Related Human Activity Recognition Using Skeletons and Multi-Stage Deep Learning for Healthcare}

\author{ \uppercase{Subrata Kumer Paul\authorrefmark{1,2}}, \uppercase{Abu Saleh Musa Miah \authorrefmark{3, 4},\IEEEmembership{Member, IEEE}}, \uppercase{Rakhi Rani Paul\authorrefmark{1,2}}, \uppercase{Md. Ekramul Hamid\authorrefmark{2}}, \uppercase{Jungpil Shin \authorrefmark{4},\IEEEmembership{Senior Member, IEEE}}, \uppercase{Md Abdur Rahim \authorrefmark{5}} }
\address[1]{Dept. of Computer Science and Engineering (CSE), Bangladesh Army University of Engineering \& Technology (BAUET), Qadirabad Cantonment, Dayarampur, Natore-6431, Rajshahi, Bangladesh.}
\address[2]{Department of Computer Science \& Engineering, University of Rajshahi, Rajshahi-6205, Bangladesh.}
\address[3]{Dept. of Computer Science and Engineering (CSE), Bangladesh Army University of Science and Technology (BAUST), Saidpur, Bangladesh.}
\address[4]{School of Computer Science and Engineering, The Univeristy of Aizu, Aizuwakmatsu, Fukushima, Japan.}
\address[5]{Dept. of Computer Science and Engineering, Pabna University of Science and Technology, Rajapur, Pabna, Bangladesh.}

\tfootnote{This work was supported by the Competitive Research Fund of The University of Aizu, Japan.}
\markboth
{....}
{This paper is currently under review for possible publication in IEEE Access.}

\corresp{Corresponding author: Md. Ekramul Hamid (e-mail: ekram\_hamid@yahoo.com ) and Jungpil Shin (e-mail: jpshin@u-aizu.ac.jp).}


\begin{abstract}
The Internet of Things (IoT) and mobile technology have significantly transformed healthcare by enabling real-time monitoring and diagnosis of patients. Recognizing Medical-Related Human Activities (MRHA) is pivotal for healthcare systems, particularly for identifying actions critical to patient well-being. However, challenges such as high computational demands, low accuracy, and limited adaptability persist in Human Motion Recognition (HMR). While some studies have integrated HMR with IoT for real-time healthcare applications, limited research has focused on recognizing MRHA as essential for effective patient monitoring. This study proposes a novel HMR method tailored for MRHA detection, leveraging multi-stage deep learning techniques integrated with IoT. The approach employs EfficientNet to extract optimized spatial features from skeleton frame sequences using seven Mobile Inverted Bottleneck Convolutions (MBConv) blocks, followed by ConvLSTM to capture spatio-temporal patterns. A classification module with global average pooling, a fully connected layer, and a dropout layer generates the final predictions. The model is evaluated on the NTU RGB+D 120 and HMDB51 datasets, focusing on MRHA such as sneezing, falling, walking, and sitting etc. It achieves 94.85\% accuracy for cross-subject evaluations and 96.45\% for cross-view evaluations on NTU RGB+D 120, along with 89.00\% accuracy on HMDB51. Additionally, the system integrates IoT capabilities using a Raspberry Pi and GSM module, delivering real-time alerts via Twilio’s SMS service to caregivers and patients. This scalable and efficient solution bridges the gap between HMR and IoT, advancing patient monitoring, improving healthcare outcomes, and reducing costs.
\end{abstract}


\titlepgskip=-15pt
\maketitle
\section{Introduction}
Human Motion Recognition (HMR) systems are designed to automatically identify and monitor individual or group activities, playing a pivotal role in healthcare by enabling the tracking of physical and medical-related activities. Among these, Medical-Related Human Activities (MRHA), such as sneezing, coughing, falling, or staggering, hold critical importance for ensuring patient safety and well-being. Despite its significance, achieving accurate detection of MRHA remains a persistent challenge in the field.
With the global elderly population projected to reach 2.1 billion by 2050, as reported by the World Health Organization (WHO) and the United Nations, the demand for effective healthcare solutions is becoming increasingly urgent  \cite{lord2006visual,egawa2023dynamic_fall_miah,hassan2024deep_har_miah,10624624_lstm_najmul_miah_har_conference}. Many elderly individuals live alone or in care facilities where the number of patients often surpasses the available healthcare professionals, resulting in insufficient monitoring and heightened risks. MRHA recognition, such as detecting falls or signs of distress, is vital in such contexts to provide real-time intervention and avoid critical incidents. Among these risks, falls represent a significant concern, with the WHO reporting approximately 646,000 deaths and over 37 million severe injuries caused by falls annually \cite{zahedian2021effect}. As the elderly population is projected to more than double by 2050, accounting for roughly 16.0\% of the global population \cite{united2021world}, the need for timely treatment and intervention in emergencies is paramount. This growing demographic trend underscores the necessity of developing automatic MRHA detection systems and other healthcare-focused HMR solutions to enhance patient safety, provide continuous monitoring, and alleviate the burden on healthcare professionals.

Medical-Related Human Activity (MRHA) recognition systems are vital for ensuring safety and providing continuous monitoring of in-home care and assisted medical facilities. These systems enable timely interventions in cases of medical emergencies, such as falls, significantly reducing mortality risk by 80\% and minimizing extended hospital stays by 26\%\cite{romeo2020image}. Current approaches for MRHA detection primarily use wearable sensors and vision-based methods\cite{hassan2024deep_har_miah,10624624_lstm_najmul_miah_har_conference}, with sensors capturing acceleration changes associated with falls~\cite{lu2018image} and vision-based systems analyzing video data~\cite{gutierrez2021comprehensive}. However, developing robust automated MRHA recognition systems is crucial to deliver timely interventions and prevent severe injuries and fatalities.

 \subsection{Current Fall Detection Systems and Their Challenges}
 Despite advancements, existing MRHA recognition technologies face significant limitations. Wearable sensor-based systems, though effective in detecting acceleration changes, often struggle with user comfort, false positives during non-critical activities, and low compliance among elderly individuals with cognitive impairments~\cite{huang2018video}. Vision-based systems provide a non-invasive alternative but raise privacy concerns, as video feeds can compromise individual anonymity and legal safeguards, even when encoding techniques are employed to obscure clarity. Skeleton-based data, derived from pose estimation algorithms~\cite{vakunov2020mediapipe} or Kinect systems~\cite{liu2020disentangling}, offers a promising solution. This approach preserves privacy by omitting identifiable information while maintaining robustness against challenges such as background noise and lighting variations. Skeleton-based data also have lower dimensionality, ensuring efficient motion representation with reduced computational costs. Recent studies have highlighted the efficacy of skeleton-based MRHA recognition. For instance, Sania et al.\cite{zahan2022sdfa} achieved over 94\% accuracy on URFD and UPFD datasets using Graph Convolutional Networks (GCNs) combined with Convolutional Neural Networks (CNNs). Similarly, Egawa et al.\cite{electronics12153234_egawa} applied a modified GCN model to the ImVia RU-Fall dataset, reporting a 99.00\% accuracy. However, these systems predominantly focus on datasets tailored for elderly individuals, such as fall detection datasets~\cite{egawa2023dynamic_fall_miah, zahan2022sdfa, li2024dmsanet}, limiting their applicability to younger populations or diverse MRHA scenarios. Moreover, existing systems are rarely evaluated on vision-based datasets designed explicitly for MRHA recognition beyond falls. Recent research has explored accelerometer-based~\cite{javeed2023deep_medical_har_acceler_signal}, audio-based~\cite{boborzi2024human_medial_har_audio_signal}, and multi-sensor approaches~\cite{boborzi2024human_medial_har_audio_signal} for MRHA recognition. Still, there is a notable lack of vision-based MRHA datasets except for those focused on elderly falls.
\subsection{Emerging Datasets and Research Gaps}
Two promising datasets, NTU RGB+D 120 and HMDB51, include medical-related human activity video data, offering new opportunities for MRHA recognition research. These datasets encompass a variety of medical activities, such as sneezing, coughing, sitting, and walking, which are relevant to healthcare scenarios. Despite this, few researchers have developed MRHA recognition systems based on these datasets, and their reported accuracy levels remain unsatisfactory.
Addressing the limitations of existing approaches and leveraging these datasets for MRHA recognition could lead to the development of more effective, privacy-preserving, and versatile systems. Future work should focus on advancing vision-based MRHA recognition models, improving their accuracy and generalizability across diverse populations and activity types.

\subsection{Motivation}
{
The increasing elderly population and the rising prevalence of falls underscore the urgent need for accurate, efficient, and privacy-preserving medical-related human activity (MRHA) recognition systems. Falls remain a leading cause of injury and mortality among older individuals, significantly impacting healthcare costs and quality of life. Traditional approaches, such as wearable sensors and vision-based systems, face challenges like low accuracy, limited portability, environmental sensitivity, and privacy concerns, leaving a gap in effective patient monitoring solutions.
This study is driven by the necessity for a robust, scalable solution to address these limitations. By integrating the Internet of Things (IoT) and mobile technology, healthcare monitoring can be transformed, enabling real-time and continuous oversight of patients' health. Focusing on skeleton data, which effectively captures joint movements while preserving privacy, this research aims to develop an advanced human motion recognition (HMR) framework tailored for healthcare applications. This approach promises to enhance safety and care for the aging population, improve patient outcomes, and reduce healthcare costs, providing a significant leap forward in modern healthcare systems.}
\subsection{The Goal and Scope of the Study}
The primary objective of this research is to develop a real-time Medical-Related Human Activity (MRHA) recognition system using a multi-stage deep learning-based spatial-temporal feature extraction technique, integrated with IoT infrastructure. By incorporating a mobile phone notification system that operates independently of third-party apps, the study aims to enable seamless communication of critical health-related events. This system is rigorously validated to ensure high accuracy and effectiveness, contributing to healthcare innovation by providing timely, data-driven insights for enhanced patient care and outcomes. The main contributions of this study are summarized as follows:

\begin{itemize}
\item \textbf{Novel Hybrid Deep Learning Model for MRHA Recognition:} We propose a multi-stage deep learning architecture for MRHA recognition, combining EfficientNet for spatial feature extraction with ConvLSTM for spatio-temporal feature integration named ENConvLSTM algorithm. This design effectively addresses challenges such as high computational demands, low accuracy, and adaptability limitations in existing Human Motion Recognition (HMR) systems. By leveraging seven Mobile Inverted Bottleneck Convolutions (MBConv) blocks in EfficientNet, the model achieves enhanced spatial feature representation and robust motion analysis.
\item \textbf{Exceptional Performance on Benchmark Datasets:} The model is evaluated on the NTU RGB+D 120 and HMDB51 datasets, focusing on MRHA such as sneezing, falling, walking, and sitting. It achieves 94.85\% accuracy for cross-subject evaluations and 96.45\% for cross-view evaluations on NTU RGB+D 120, along with 89.22\% accuracy on HMDB51. These results demonstrate the model's capability to handle both spatial and temporal data aspects effectively.
\item \textbf{Real-Time IoT-Integrated MRHA Recognition System:} A real-time IoT-based system is developed, integrating a Raspberry Pi and GSM module to deliver SMS notifications via Twilio’s API. This approach ensures instant alerts for caregivers and patients without relying on third-party applications, enhancing usability and scalability. By addressing 12 MRHAs, including activities like sneezing, falling, and walking, the system provides a proactive solution for patient monitoring, early diagnosis, and timely intervention, ultimately improving safety, and healthcare outcomes, and reducing costs.

\end{itemize}


The rest of the paper has been organized into these sections. The second section contains the Literature Review. The Major Contributions are included in Section 2. The dataset description, its selection criteria, and the preprocessing part have been mentioned in Section 4. The fifth section presents the proposed methodology. The experimental results and performance evaluation are presented in the sixth section of this paper. The real-time implementation and its analysis have been presented in the seventh section. We have discussed the limitations and future scope of the proposed system in Section 8. Finally, the paper has been concluded, and the future direction of this research is in the ninth section. The remaining parts include acknowledgment, author contributions, conflicts of interest, data availability statement, list of abbreviations, and references.

\begin{figure}[ht]
\centering
\includegraphics[scale=0.15]{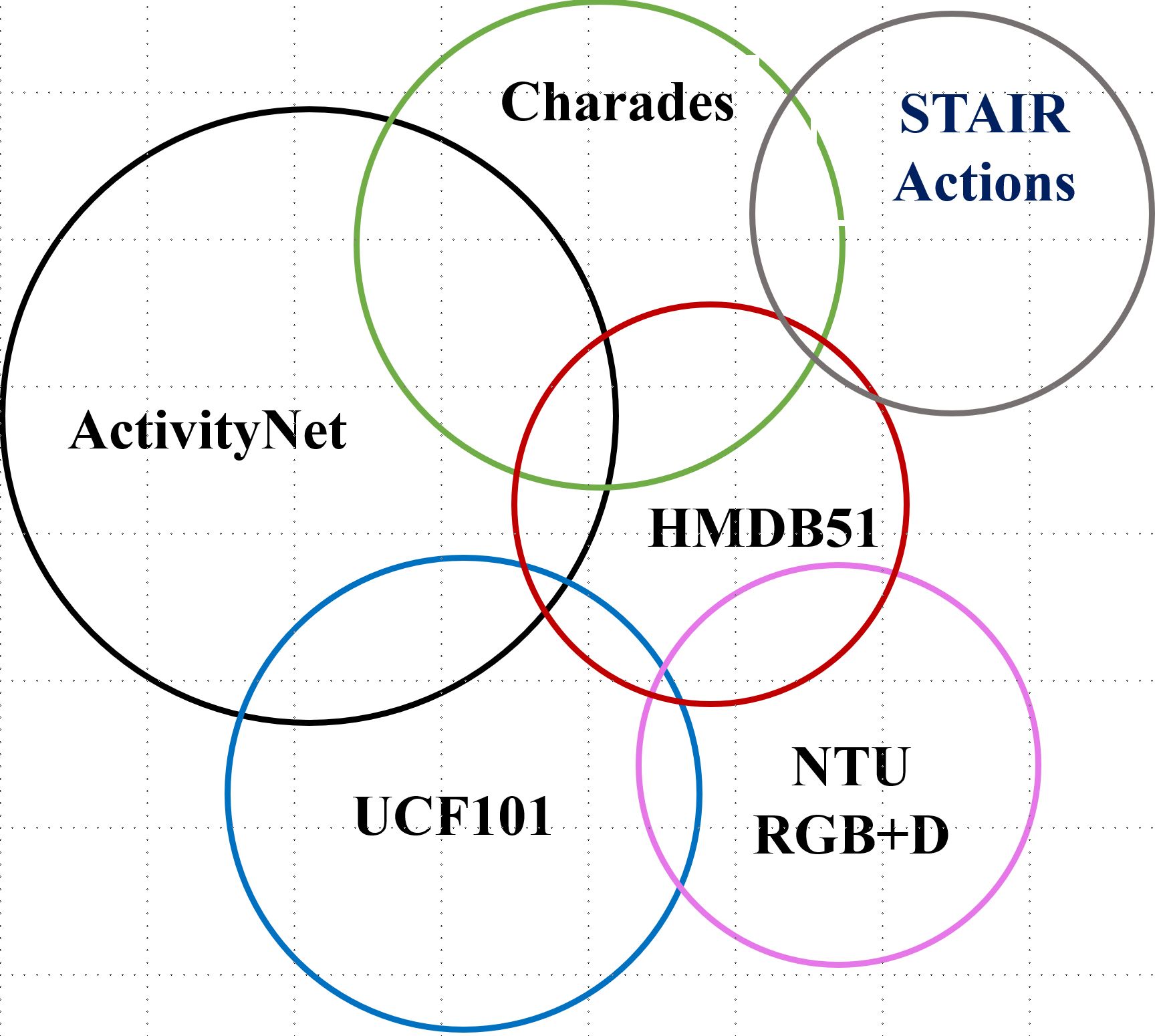}
\caption{Class overlapping among experimenting datasets. 
}
\label{fig:dataset_diagram}
\end{figure}

\begin{figure}[ht]
\centering
\includegraphics[scale=0.45]{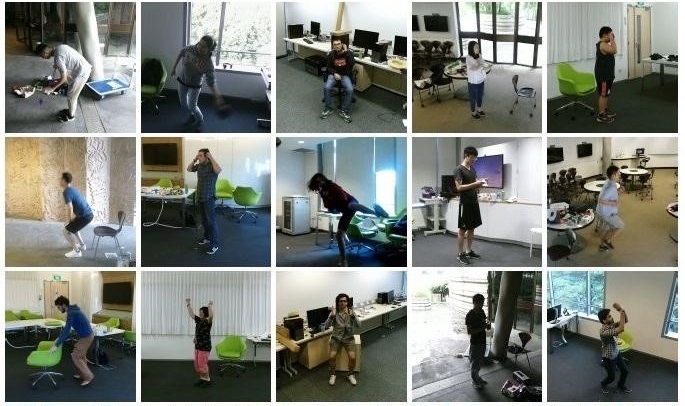}
\caption{Sample Example of "NTU RGB+D 120" Dataset.
}
\label{fig:dataset_example}
\end{figure} 

\begin{table*}[!htp]
\centering
\caption{Summarized existing works with their used techniques, results, and accuracy} \label{tab:literature_revew_summary}
\begin{tabular}{|p{3cm}|p{5cm}|p{5cm}|p{1cm}|p{1cm}|}
\hline
\textbf{Citation} & \textbf{Techniques} & \textbf{Results} & \textbf{Cross-Subject (CS)} & \textbf{Cross-View (CV)} \\
\hline
Heidari and Iosifidis, 2021 \cite{heidari2021progressive} & Progressive Spatio-Temporal Graph Convolutional Network (PST-GCN) & Combined GCN and transformer architecture for HAR. & 89.00\% & 95.00\% \\
Geng et al., 2024 \cite{geng2024hierarchical} & Hierarchical Aggregation GCN & Aggregated local and global action representations for better recognition. & 87.50\% & 93.90\% \\
Tu et al., 2023 \cite{tu2023joint} & Joint-Bone Fusion Graph Convolutional Network (J-BGCN) & Modeled joint and bone relations to improve performance. & 85.50\% & 92.60\% \\
Shi et al., 2018 \cite{shi2018adaptive} & Adaptive Skeleton Graph Network (AdaSGN) & Adaptively learned skeleton graph for more flexible representations. & 88.50\% & 94.20\% \\
Plizzari et al., 2021 \cite{plizzari2021spatial} & Spatial-Temporal Transformer Network (ST-TR) & Combined GCN with a transformer for global attention in HAR. & 88.60\% & 94.70\% \\
Song et al., 2022 \cite{song2022temporal} & Temporal-Aware Adaptive Skeleton Graph Network (TA-ASGN) & Modeled temporal adaptive skeleton graphs with transformers. & 89.80\% & 95.30\% \\
Duan et al., 2022 \cite{duan2022hybrid} & Graph Convolution and Transformer Hybrid Model & Hybrid architecture for spatio-temporal skeleton action recognition. & 90.10\% & 96.20\% \\
Zhao et al., 2022 \cite{zhao2022learning} & Skeleton-Aware Geometry Feature Learning & Proposed geometry-based skeleton features with improved accuracy. & 90.00\% & 95.80\% \\
Yang et al., 2023 \cite{yang2023temporal} & Temporal Edge Aggregation for GCN & Temporal edge aggregation for enhanced skeleton data recognition. & 90.30\% & 96.00\% \\
Feng et al., 2023 \cite{feng2023graph} & Graph Attention Network + Transformer & Attention mechanism for skeleton action recognition. & 89.70\% & 95.70\% \\
Wu et al., 2024 \cite{wu2024dual} & Dual Stream Transformer GCN & Proposed dual stream model combining temporal and spatial streams. & 91.00\% & 96.50\% \\
\hline
\end{tabular}
\end{table*}
\section{Related Work}
\label {sec:Literature Review} 
The convergence of IoT and mobile technologies has revolutionized healthcare, enabling real-time patient monitoring and diagnosis. One of the crucial areas in this domain is human motion recognition (HMR), which various methodologies have employed over the last few years. In this survey, we consider only the ``NTU RGB+D 120" dataset. Cross-Subject (CS) and Cross-View (CV) are two accuracy evaluation methods. The following Table \ref{tab:literature_revew_summary} presents the existing works with their achieved accuracy:

Peng et al. (2021) \cite{heidari2021progressive} proposed a novel architecture for human activity recognition (HAR) that integrates a transformer model with a Progressive Spatio-Temporal Graph Convolutional Network (PST-GCN). This approach effectively captures both spatial and temporal dependencies within activity sequences, combining the strengths of both GCN and transformer techniques. By applying this architecture, the model achieved impressive results, with a cross-subject accuracy of 89.00\% and a cross-view accuracy of 95.00\%. These results highlight the model's robustness in handling both subject variations and changes in viewpoint, which are critical challenges in HAR systems.

Chen et al. (2021) \cite{geng2024hierarchical} introduced a Hierarchical Aggregation Graph Convolutional Network (GCN) designed to improve human activity recognition (HAR) by effectively combining local and global action representations. This hierarchical structure enhances the model's ability to capture complex interactions between different body parts and activities. The approach demonstrated strong performance, achieving 87.50\% cross-subject accuracy and 93.90\% cross-view accuracy, underscoring its effectiveness in generalizing across diverse subjects and viewpoints in HAR tasks. Tu et al. (2023) \cite{tu2023joint} developed the Joint-Bone Fusion Graph Convolutional Network (J-BGCN), which focuses on enhancing human activity recognition (HAR) by modeling both joint and bone relationships. By fusing these two key aspects of motion data, the network captures richer spatial information, leading to improved recognition performance. The J-BGCN achieved 85.50\% cross-subject accuracy and 92.60\% cross-view accuracy, demonstrating its ability to effectively learn from joint-bone interactions and improve the overall robustness of HAR systems. Shi et al. (2018) \cite{shi2018adaptive} introduced the Adaptive Skeleton Graph Network (AdaSGN), which adaptively learns the structure of skeleton graphs to provide more flexible and dynamic representations for human activity recognition (HAR). This adaptive approach allows the model to capture the nuances of human movement better, enhancing its ability to generalize across various activities. AdaSGN achieved a cross-subject accuracy of 88.50\% and a cross-view accuracy of 94.20\%, demonstrating its effectiveness in handling diverse subject conditions and viewpoints in HAR tasks.
Plizzari et al. (2021) \cite{plizzari2021spatial} proposed the Spatial-Temporal Transformer Network (ST-TR), which integrates a Graph Convolutional Network (GCN) with a transformer to enhance human activity recognition (HAR) by capturing global attention across spatial and temporal dimensions. This hybrid model leverages the GCN's ability to model local dependencies and the transformer's strength in capturing long-range interactions. The ST-TR achieved a cross-subject accuracy of 88.60\% and a cross-view accuracy of 94.70\%, showcasing its robustness in addressing both spatial and temporal complexities in HAR tasks.
Song et al. (2022) \cite{song2022temporal} introduced the Temporal-Aware Adaptive Skeleton Graph Network (TA-ASGN), a model that combines transformers with temporal adaptive skeleton graphs for improved human activity recognition (HAR). This approach enhances the flexibility of the skeleton graph while incorporating temporal dependencies through the transformer, allowing for more precise modeling of complex human movements. TA-ASGN achieved a cross-subject accuracy of 89.80\% and a cross-view accuracy of 95.30\%, demonstrating superior performance in both subject variation and viewpoint adaptation in HAR tasks.
Duan et al. (2022) \cite{duan2022hybrid} developed a Graph Convolution and Transformer Hybrid Model designed for spatio-temporal skeleton-based action recognition in human activity recognition (HAR). By integrating the strengths of graph convolutional networks (GCNs) for local spatial feature extraction and transformers for global temporal attention, the model effectively captures complex activity patterns. This hybrid approach achieved impressive results, with a cross-subject accuracy of 90.10\% and a cross-view accuracy of 96.20\%, highlighting its robustness and effectiveness in handling both spatial and temporal aspects of skeleton-based action recognition.  Zhao et al. (2022) \cite{zhao2022learning} introduced a novel approach to human activity recognition (HAR) through Skeleton-Aware Geometry Feature Learning, which leverages geometry-based skeleton features to enhance recognition accuracy. By focusing on the geometric relationships within the skeleton data, the model can capture more detailed movement patterns, leading to improved performance. The method achieved a cross-subject accuracy of 90.00\% and a cross-view accuracy of 95.80\%, demonstrating its effectiveness in refining skeleton-based HAR through geometry-aware feature learning. Yang et al. (2023) \cite{yang2023temporal} proposed an innovative approach to human activity recognition (HAR) by introducing Temporal Edge Aggregation for Graph Convolutional Networks (GCN). This technique focuses on enhancing the recognition of skeleton data by aggregating temporal edge information, which improves the model's ability to capture dynamic and sequential aspects of human movements. The method achieved a cross-subject accuracy of 90.30\% and a cross-view accuracy of 96.00\%, highlighting its effectiveness in improving the temporal modeling and recognition capabilities of GCN-based HAR systems.

Feng et al. (2023) \cite{feng2023graph} introduced a hybrid model that combines a Graph Attention Network (GCN) with a transformer to enhance skeleton-based human activity recognition (HAR). The model can selectively focus on important spatial-temporal features in skeleton action data by incorporating an attention mechanism, leading to more accurate recognition. This approach achieved a cross-subject accuracy of 89.70\% and a cross-view accuracy of 95.70\%, demonstrating its effectiveness in capturing local and global dependencies for improved performance in HAR tasks.

Wu et al. (2024) \cite{wu2024dual} proposed the Dual Stream Transformer GCN, a sophisticated model that integrates both temporal and spatial streams for enhanced human activity recognition (HAR). This dual-stream approach allows the model to simultaneously capture dynamic temporal changes and spatial relationships within the data, resulting in a more comprehensive understanding of human activities. The Dual Stream Transformer GCN achieved a cross-subject accuracy of 91.00\% and a cross-view accuracy of 96.50\%, underscoring its superior performance in addressing the complexities of HAR tasks through the effective combination of temporal and spatial features.

Sania et al.\cite{zahan2022sdfa} achieved over 94\% accuracy on URFD and UPFD datasets using Graph Convolutional Networks (GCNs) combined with Convolutional Neural Networks (CNNs). Similarly, Egawa et al.\cite{electronics12153234_egawa} applied a modified GCN model to the ImVia RU-Fall dataset, reporting a 99.00\% accuracy. However, these systems predominantly focus on datasets tailored for elderly individuals, such as fall detection datasets~\cite{egawa2023dynamic_fall_miah, zahan2022sdfa, li2024dmsanet}, limiting their applicability to younger populations or diverse MRHA scenarios. Moreover, existing systems are rarely evaluated on vision-based datasets designed explicitly for MRHA recognition beyond falls. Recent research has explored accelerometer-based~\cite{javeed2023deep_medical_har_acceler_signal}, audio-based~\cite{boborzi2024human_medial_har_audio_signal}, and multi-sensor approaches~\cite{boborzi2024human_medial_har_audio_signal} for MRHA recognition. Still, there is a notable lack of vision-based MRHA datasets except for those focused on elderly falls. Two promising datasets, NTU RGB+D 120 and HMDB51, include medical-related human activity video data, offering new opportunities for MRHA recognition research. These datasets encompass a variety of medical activities, such as sneezing, coughing, sitting, and walking, which are relevant to healthcare scenarios. Despite this, few researchers have developed MRHA recognition systems based on these datasets, and their reported accuracy levels remain unsatisfactory. Addressing the limitations of existing approaches and leveraging these datasets for MRHA recognition could lead to the development of more effective, privacy-preserving, and versatile systems. Future work should focus on advancing vision-based MRHA recognition models, improving their accuracy and generalizability across diverse populations and activity types.

\section{Dataset}
This paper explores the use of existing HMR datasets instead of selecting a new one. Three key criteria guided dataset selection: (1) availability of activities listed in Table II, (2) relevance of selected activities to the medical domain, and (3) richness of features in the video dataset. Six video datasets related to HMR were analyzed, as detailed in Table \ref{tab:hmr_dataset_information}. Figure \ref{fig:dataset_diagram} illustrates overlapping classes among these datasets. The ActivityNet and Charades datasets share the highest number of overlapping classes, indicating a strong correlation with the “NTU RGB+D 120” dataset. The STAIR Actions dataset also shares some common classes, while UCF101 and HMDB51 have the least overlap, suggesting a weak correlation with “NTU RGB+D 120.”
Based on these observations, we selected the “NTU RGB+D 120” dataset for our experiments. Additionally, its current accuracy levels, as discussed in the literature, remain insufficient, further motivating its use in our study \url{https://rose1.ntu.edu.sg/dataset/actionRecognition/}. 

\begin{table}[!htp]
\centering
\caption{Human Motion Recognition (HMR) datasets with some class activities} \label{tab:hmr_dataset_information}
\begin{tabular}{|c|l|c|c|}
\hline
\textbf{SN} & \textbf{Dataset Name} & \textbf{Total Samples Data} & \textbf{Total Classes} \\ \hline
1 & ActivityNet          & 21,313  & 200  \\ \hline
2 & Charades             & 66,493  & 157  \\ \hline
3 & HMDB51               & 6,766   & 51   \\ \hline
4 & NTU RGB+D 120       & 114,480 & 120  \\ \hline
5 & STAIR Actions        & 109,478 & 100  \\ \hline
6 & UCF101               & 13,320  & 101  \\ \hline
\end{tabular}
\end{table}

\subsection{NTU RGB+D 120 dataset}
The “NTU RGB+D 120” dataset, developed by Nanyang Technological University, serves as a widely recognized benchmark for human action recognition research. This large-scale dataset includes 120 distinct human motion classes, totaling 114,480 samples, with RGB video data (.mp4) captured at a resolution of 1920x1080 pixels (aspect ratio 16:9) and a frame rate of 24 fps. Additionally, each sample provides 3D coordinates of 25 body joints per frame, enabling detailed skeletal motion analysis that is presented in Figure \ref{fig:keypoints}. For this study, we focus on 12 specific medical-related activity classes in .mp4 format, encompassing sneeze/cough (950 samples), staggering (1050), chest pain (1100), back pain (1150), fan self (850), yawn (950), falling (1000), headache (900), neck pain (950), vomiting (1000), stretch oneself (1200), and blow nose (1100). These activities amount to a total of 13,200 samples, specifically curated to address challenges in healthcare-related applications. The dataset is divided into 80\% training (10,560 samples) and 20\% testing (2,640 samples) splits, ensuring robust evaluation through cross-subject and cross-view experimental setups. This splitting strategy maintains a balanced coverage of classes, supporting comprehensive model validation. Table \ref{tab:dataset_details} summarizes the selected medical classes along with their respective sample sizes, while Figure \ref{fig:dataset_example} provides visual examples from the dataset.

Researchers typically use the “NTU RGB+D 120” dataset in two configurations, described below:
\begin{itemize} \item \textbf{Cross-subject:} This configuration evaluates the model performance by training on data from certain subjects and testing on data from different subjects, assessing generalization across individuals. \item \textbf{Cross-view:} This configuration trains the model on data from one viewpoint (e.g., frontal view) and tests it on data from another viewpoint (e.g., side view), evaluating its adaptability to different perspectives. \end{itemize}
In our experiments, we focus on 12 medical-related classes from the “NTU RGB+D 120” dataset, representing common healthcare scenarios. Figure \ref{fig:dataset_diagram} illustrates sample frames from this dataset.

\subsection{HMDB51 Dataset}
We also conducted experiments using the HMDB51 dataset, a well-established benchmark in human action recognition research. The HMDB51 dataset comprises 6,766 video clips with a total file size of 2GB. It features 51 action categories, with each category containing at least 101 video clips sourced from diverse origins, including movies, YouTube, and other online platforms. For this study, we selected six specific classes—floor, walk, stand, eat, sit, and drink—that are particularly relevant to medical and healthcare applications. These classes were chosen due to their critical importance in healthcare scenarios where activity recognition can provide meaningful insights and enhance patient monitoring systems. Table \ref{tab:hmr_dataset_information} provides detailed information about the HMDB51 dataset and its class-wise distribution, focusing on the six selected medical-related activities. This selection aims to align the dataset's versatility with the specific requirements of healthcare-related human activity recognition tasks.
\begin{table}[]
\caption{The 12 medical classes with their corresponding sample sizes (including data splitting)} \label{tab:dataset_details}
\begin{tabular}{|p{3cm}|p{1cm}|p{2cm}|p{1cm}|}
\hline
Action Name& No. of skeletal samples & Action Name & No. of skeletal samples \\ \hline
A41:   sneeze/cough & 950 & A42:   staggering & 1050 \\ \hline
A45: chest pain  & 1100 & A46:   back pain & 1150 \\ \hline
A49: fan   self & 850 & A103:   yawn & 950 \\
A43:   falling down & 1000 & A44: headache & 900 \\ \hline
A47:   neck pain & 950 & A48: vomiting & 1000 \\ \hline
A104:   stretch oneself & 1200 & A105:   blow nose & 1100 \\ \hline
\multirow{2}{*}{Total   No. of samples} & 6050 &  & 6150 \\ 
 & \multicolumn{3}{|p{4cm}|}{13,200} \\ \hline
Cross-Subject and Cross-View Evaluation Methods
(80\% Training and 20\% 
Testing Split)
 &  \multicolumn{3}{|p{4cm}|}{Training Set: 80\% of 13,200 = 10,560 samples,
Testing Set: 20\% of 13,200 = 2,640 samples
}
\\ \hline
\end{tabular}
\end{table}

\begin{figure*}[ht]
\centering
\includegraphics[scale=0.23]{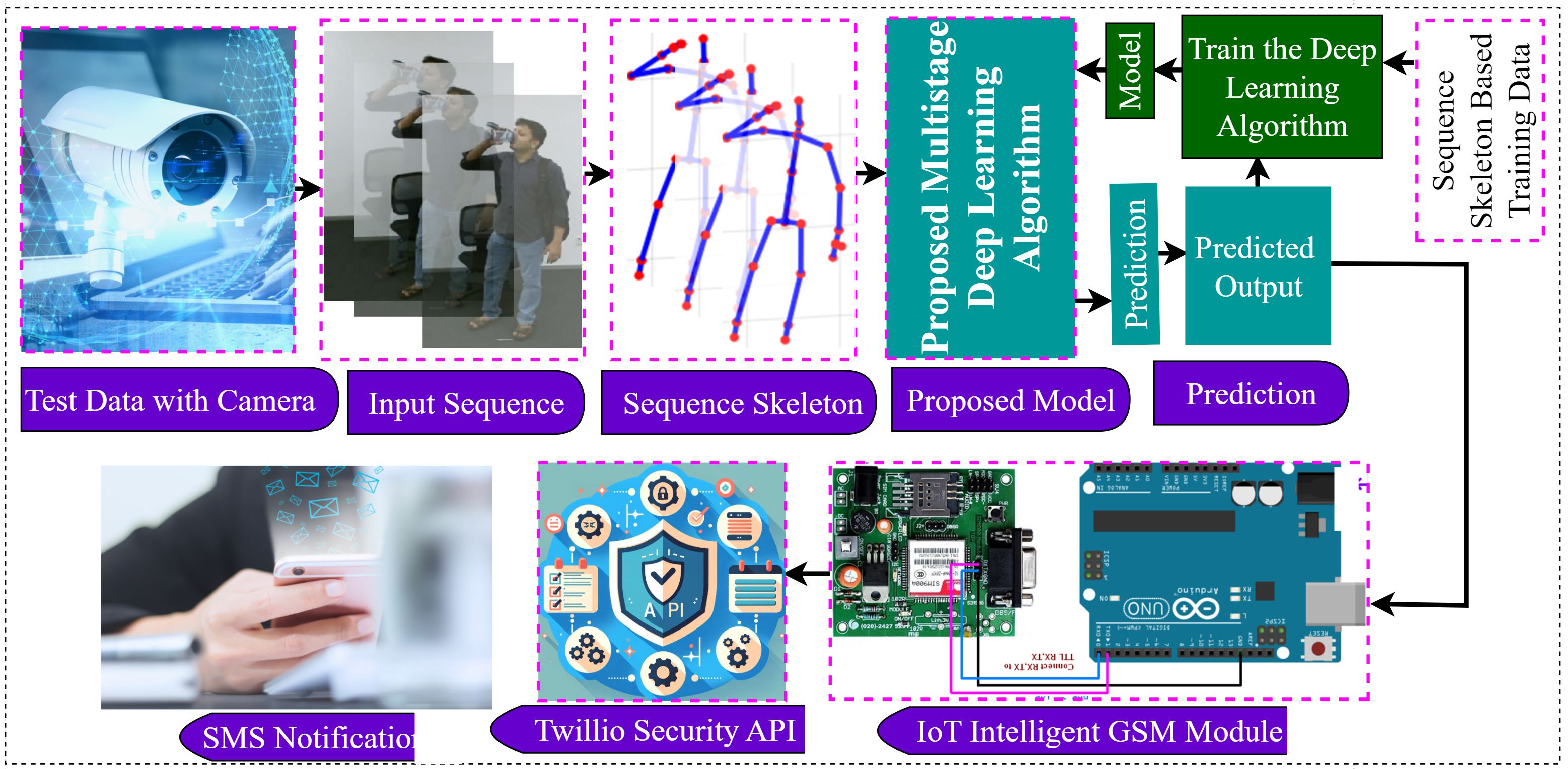}
\caption{The workflow of the proposed methodology this figure represents an end-to-end system for human 
motion detection using a deep learning model, followed by real-time monitoring and notification. 
}
\label{fig:proposed_main_diagram}
\end{figure*} 


\begin{figure*}[ht]
\centering
\includegraphics[scale=0.30]{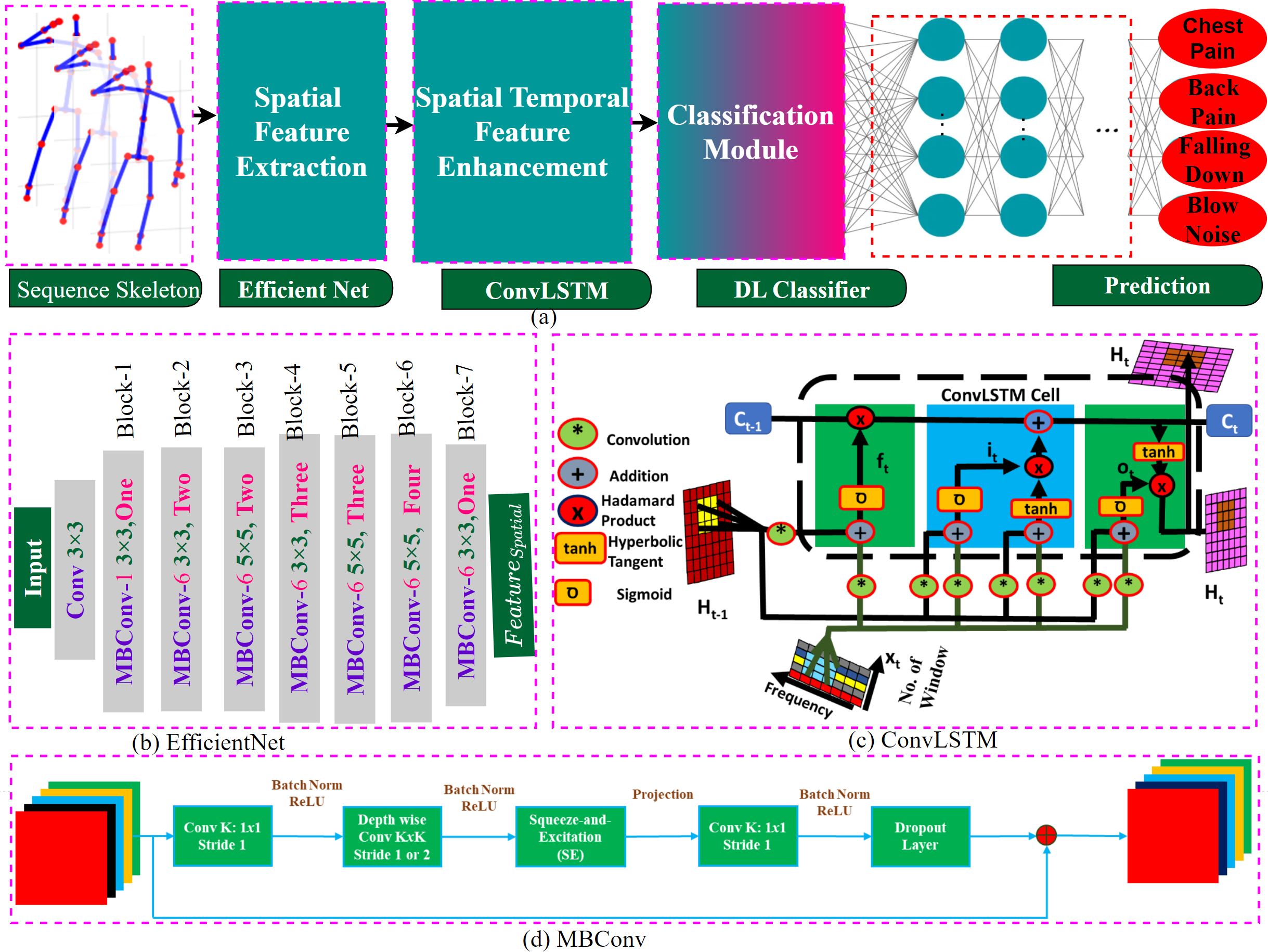}
\caption{ (a) Proposed multi-stage deep learning model constructed with (b) EfficientNet and (C) ConvLSTM beside the classification module (d) MB Convolution. 
}
\label{fig:proposed_dl_diagram}
\end{figure*} 
\section{Propsoed Method}
\label {sec: Materials and methods}
There are many researchers have been working to develop a human motion recognition (HMR) system. However, a few researchers have been working to develop IoT integrating HMR systems. we propose a new HMR method that uses spatial and temporal features powered by multi-stage deep learning and integrated with IoT. First, we use EfficientNet to extract spatial features from skeleton frame sequences. EfficientNet is designed with seven Mobile Inverted Bottleneck Convolutions (MBConv) blocks. Each block includes a convolutional layer, a depthwise separable layer, and a squeeze-and-excitation (SE) module to create optimized feature representations. These spatial features are then passed to ConvLSTM, which extracts spatio-temporal features, combining spatial and sequential information. The main components of our study are given below:  
\begin{itemize}
\item 
 \textbf{OpenPose Based BodyPose Extraction:} We employed OpenPose to extract 25 key points from the whole for each frame in the sequence. 
 \item 
\textbf{Hybrid Deep Learning Architecture}
We introduce a novel hybrid deep learning model, ENConvLSTM, combining EfficientNet and ConvLSTM to address challenges in HMR, such as high computational demands, low accuracy, and adaptability. The proposed architecture consists of two key components, efficientNet to extract the spatial feature from 
input skeleton data. It utilizes seven Mobile Inverted Bottleneck Convolutions (MBConv) blocks. Each MBConv block comprises a 1x1 convolution layer for feature mapping a depthwise separable convolution layer for dimensionality reduction. A squeeze-and-excitation (SE) module for adaptive feature recalibration. Then we extracted the spatio-temporal feature within ConvLSTM. This takes the spatial features generated by EfficientNet and models the temporal dependencies across frames, producing robust spatio-temporal features.
\item
\textbf{Classification Module:}
The spatio-temporal features generated by ConvLSTM are passed through a classification module, which includes Global Average Pooling and Fully Connected Layer etc.  This multi-stage pipeline ensures robust motion recognition by integrating spatial and temporal modeling techniques.
\item 
\textbf{IoT Integration for Real-Time Alerts:} 
The system integrates IoT components, including a Raspberry Pi and GSM module, to provide real-time alerts. Twilio’s SMS API is used to send instant notifications to caregivers and patients, removing the dependency on third-party mobile applications. This feature enhances the system's scalability and usability for healthcare scenarios.
\end{itemize}

\begin{figure}[ht]
\centering
\includegraphics[scale=0.15]{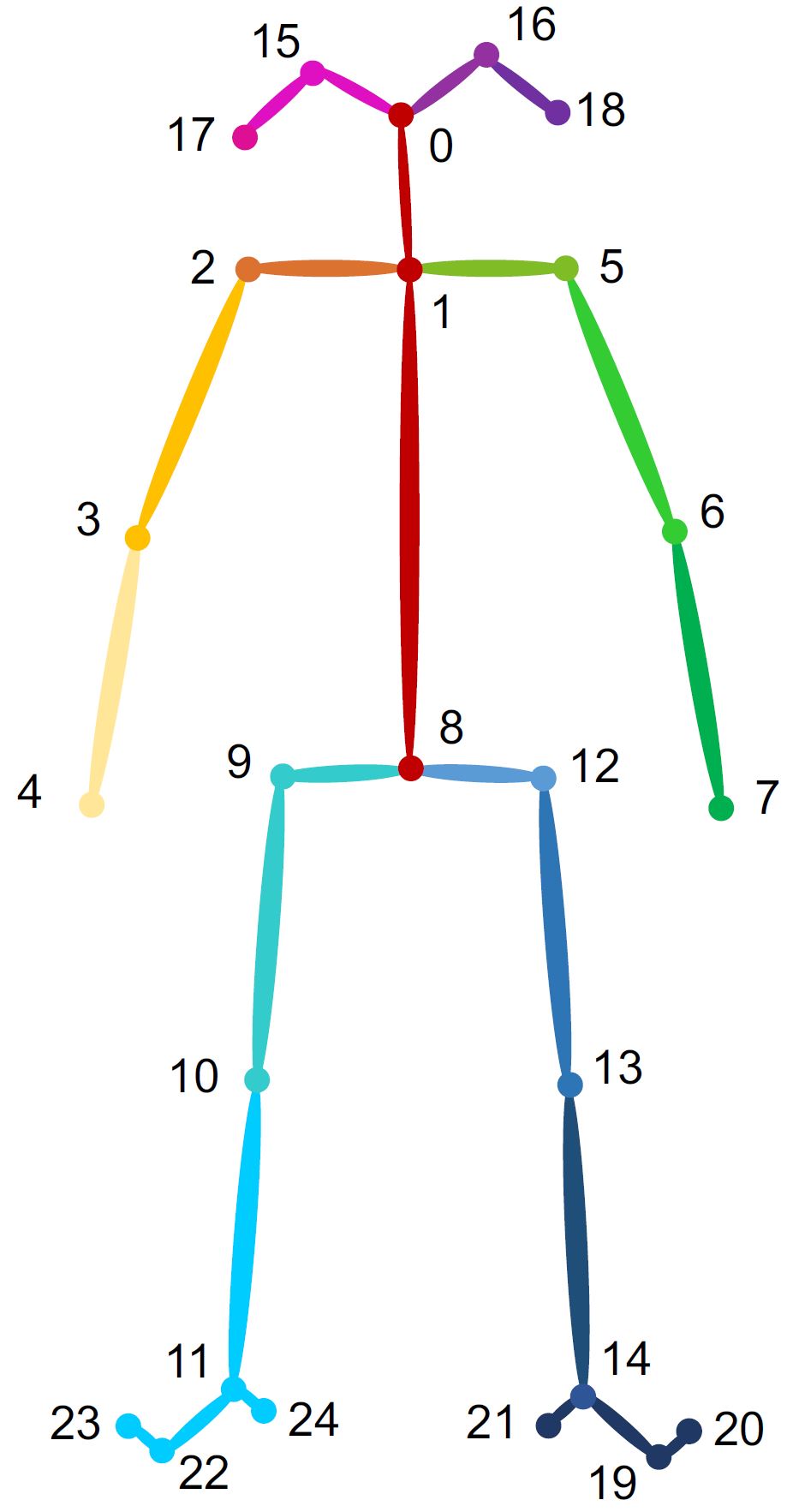}
\caption{Illustration of the configuration of 25 body joints in our dataset (a). The labels of these joints are (b): (1) base of spine, (2) middle of the spine, (3) neck, (4) head, (5) left shoulder, (6) left elbow, (7) left wrist, (8) left hand, (9) right shoulder, (10) right elbow, (11) right wrist, (12) right hand, (13) left hip, (14) left knee, (15) left ankle, (16) left foot, (17) right hip, (18) right knee, (19) right ankle, (20) right foot, (21) spine, (22) tip of the left hand, (23) left thumb, (24) tip of right hand, (25) right thumb
\cite{zhang2023multi_skeleton25_points}.}
\label{fig:keypoints}
\end{figure}
The data preprocessing and each step of the feature extraction with real-time implementation IoT integration are described below. 

\subsection{Data Preprocessing }The preprocessing and pose keypoint extraction process begins with detecting 25 skeletal joints of the human body using the Kinect v2 camera. These joints include key points such as the head, shoulders, elbows, wrists, hips, knees, ankles, and feet, each represented by 3D coordinates (X, Y, Z). RGB videos are recorded at a resolution of 1920×1080, while depth maps are captured at 512×424 resolution. Skeletons are extracted from video frames at a rate of 24 frames per second (FPS), enabling precise tracking of motion trajectories. Figures \ref{fig:proposed_main_diagram} and \ref{fig:keypoints} illustrate the skeletal configuration, which abstracts human poses and movements while preserving spatial and temporal features crucial for motion analysis.

The OpenPose library is used to extract skeletal data, identifying and tracking the 25 body joints for each individual in the scene. Each frame provides a reduced yet comprehensive skeletal representation of human movement, significantly simplifying raw video data. This abstraction captures essential motion patterns, allowing for the analysis of complex motion dynamics while reducing computational complexity. The skeleton data structure provides an efficient input format for deep learning models, focusing on critical movement patterns.

The preprocessing pipeline further enhances the data for analysis. Video frames are sampled at 10 frames per second (FPS) to eliminate redundancy while retaining critical motion information. Each frame is resized to a uniform resolution, converted to grayscale, and normalized to ensure consistency and compatibility across samples. Finally, the processed frames are organized into sequential arrays to represent the temporal dynamics of motion and fed into the feature extraction module. 
\subsection{Spatial Temporal Feature Extraction}
The sequential skeleton information is fed into the feature extraction module. here first we extract the spatial feature using EfficientNet and then we fed him into ConvLSTM to extract the spatiotemporal features defined below. 


\subsubsection{EfficientNet}
EfficientNet is a state-of-the-art deep learning model for spatial feature enhancement and image classification tasks. It achieves high accuracy with fewer parameters and lower computational costs by scaling depth, width, and resolution in a balanced manner due to depthwise separable convolution and sequence excitation module. Figure \ref{fig:proposed_dl_diagram}(b) shows the efficient net Model diagram, which was constructed with various deep learning modules to extract the spatial feature from input skeleton data. It utilizes seven Mobile Inverted Bottleneck Convolutions (MBConv) blocks. Each MBConv block comprises a 1x1 convolution layer for feature mapping and a depthwise separable convolution layer for dimensionality reduction. A squeeze-and-excitation (SE) module for adaptive feature recalibration. 
The series of MBConv is mainly used to downsample and extract meaningful features from the input which is demonstrated in Figure \ref{fig:proposed_dl_diagram} (d). The structure consists of multiple blocks of convolutions, where the operations can be represented as:
\begin{equation}
y = f(x; W) = \text{MBConv}(x; W)
\label{eq:mbconv}
\end{equation}

Where \(x\) is the input skeleton features, \(W\) are the convolutional weights, and \(y\) is the output feature map. EfficientNet progressively reduces the spatial resolution while expanding the depth of features, leading to high-level representations for the next stage. These blocks apply depth-wise separable convolutions to efficiently extract hierarchical spatial features, resulting in feature maps from progressively lower resolutions and deeper feature representations.

\subsubsection{ConvLSTM}
Then we extracted the spatio-temporal feature within ConvLSTM. This takes the spatial features generated by EfficientNet and models the temporal dependencies across frames, producing robust spatio-temporal features. This hybrid approach addresses the limitations of existing methods, such as high computational complexity, low accuracy, and limited adaptability to diverse healthcare scenarios. Figure \ref{fig:proposed_dl_diagram}(c) demonstrated the  ConvLSTM networks diagram which mainly extended the capabilities of traditional LSTM networks by integrating convolutional layers. This structure allows ConvLSTM to handle spatial data effectively, making it well-suited for sequential data with spatial dependencies, such as video frames or skeleton sequences. ConvLSTM captures the temporal dynamics of human motion, which is crucial for accurate recognition.
The output of EfficientNet acts as a feature extractor, feeding into the ConvLSTM module shown at the bottom. The ConvLSTM, represented by its internal gates, handles temporal dynamics, crucial for modeling time-sequenced data. The core equations governing ConvLSTM include the forget gate, input gate, cell state update, and output gate, which are mathematically expressed using sigmoid (\(\sigma\)) and tanh activations to manage memory and cell state transitions.

\begin{equation}
f_t = \sigma(W_x^f * X_t + W_h^f * H_{t-1} + b_f)
\label{eq:forget_gate}
\end{equation}

\begin{equation}
i_t = \sigma(W_x^i * X_t + W_h^i * H_{t-1} + b_i)
\label{eq:input_gate}
\end{equation}

\begin{equation}
C_t = f_t \odot C_{t-1} + i_t \odot \tanh(W_x^c * X_t + W_h^c * H_{t-1} + b_c)
\label{eq:cell_state_update}
\end{equation}

\begin{equation}
o_t = \sigma(W_x^o * X_t + W_h^o * H_{t-1} + b_o)
\label{eq:output_gate}
\end{equation}

\begin{equation}
H_t = o_t \odot \tanh(C_t)
\label{eq:hidden_state}
\end{equation}

The symbols used in the ConvLSTM equations represent key components of its operation: \(f_t\), \(i_t\), and \(o_t\) denote the forget, input, and output gate activations at time step \(t\), while \(C_t\) and \(H_t\) represent the cell state and hidden state, respectively. The sigmoid (\(\sigma\)) and hyperbolic tangent (\(\tanh\)) functions are used as activation functions. \(W_x^f, W_x^i, W_x^c, W_x^o\) and \(W_h^f, W_h^i, W_h^c, W_h^o\) are weight matrices for the input \(X_t\) and hidden state \(H_{t-1}\), respectively, in each gate, with \(b_f, b_i, b_c, b_o\) as their corresponding bias terms. The Hadamard product (\(\odot\)) represents element-wise multiplication. Information flows through these gates, where the input (\(X_t\)) and previous hidden state (\(H_{t-1}\)) are selectively remembered or forgotten over time. This architecture outputs temporal features, classified into predefined activity classes (e.g., sneezing, yawning, or back pain from A41–A105 classes) using a SoftMax layer, as shown in Figure \ref{fig:proposed_dl_diagram}. By capturing spatial and temporal dependencies, this approach proves highly effective for human motion recognition tasks.

\subsection{Classification Module}
The spatio-temporal features generated by ConvLSTM are passed through a classification module, which includes Global Average Pooling and Fully Connected Layer etc.  This multi-stage pipeline ensures robust motion recognition by integrating spatial and temporal modeling techniques.
 The output of the classification model represents the classified activity or motion. The classification loss function \( L \) is calculated using the following equation:
\begin{equation}
L = -\sum_{i=1}^{N} y_i \cdot \log(\hat{y}_i)
\end{equation}
 Where \( y_i \) is the true label, \( \hat{y}_i \) is the predicted probability from the SoftMax output, and \( N \) is the number of classes. After the model is trained and evaluated, it can be deployed for real-time human motion detection. A CCTV Camera is used to capture live video input, which is then processed by the system to detect human motions (such as chest pain, headache, vomiting, etc.). In the real-time monitoring system, the model outputs the detected human motion. This output is fed into an IoT Intelligent GSM Module (SIM900A GSM Module), which is responsible for communicating with a security API. This module allows the system to send notifications or alerts based on detected activities.

\section{Experiment and Result}
The proposed model is evaluated on the "NTU RGB+D 120" dataset, focusing on 12 selected medical classes. The dataset is divided into training (80\%) and testing (20\%) portions. The model is trained on 80\% of the data, with its performance evaluated on the remaining 20\%. During training, the ENConvLSTM model minimizes classification loss, typically cross-entropy, using the Adam optimizer to learn and differentiate between various human activities. Key performance metrics—accuracy, precision, recall, and F1-score—are calculated for both cross-subject and cross-view evaluations \cite{miah2017motor_miah_SVM_PCA_ANOVA,ali2022potential_miah,miah2020motor_pca_Anova_LDA_mad,hossain2023exploring_miah}. The model performs exceptionally well, particularly in cross-view evaluations, which involve greater variability due to different camera angles, demonstrating its robustness and generalizability for real-time Human Motion Recognition (HMR) in healthcare applications. To further validate its effectiveness, the ENConvLSTM model’s results are compared with state-of-the-art methods, showcasing significant improvements. 

These metrics comprehensively evaluate the model’s performance in recognizing human motion and underline the ENConvLSTM model’s effectiveness for healthcare applications.

\subsection{Software and Hardware Requirements}
The study requires several key software and hardware components. The software includes deep learning frameworks like TensorFlow, PyTorch, scikit-learn, and Keras, with data processing libraries such as NumPy, Pandas, and OpenCV, all using Python. Development is done in Jupyter Notebook and PyCharm. The hardware setup features an AMD Ryzen 9 5900X 12-Core Processor, running on a 64-bit system with Python 3.9.13 and CUDA 11.0. It includes an NVIDIA® GeForce RTX 3060 graphics card with 6 GB of memory, 64 GB of RAM, and a 4 TB SSD for storage. An Arduino UNO REV3 and a SIM900A GSM module are also used for SMS communication, ensuring a compact and versatile integration of hardware and software for deep learning and data processing.
\begin{table}[!htp]
\centering
\caption{Prposed model parameters} \label{tab:par_enconvlstm}
\begin{tabular}{|c|l|l|}
\hline
\textbf{SN} & \textbf{Model Parameters}       & \textbf{Value Count and Function}   \\ \hline
1           & Total Parameter                 & 8,138,104                          \\ \hline
2           & Trainable Parameter             & 8,095,640                          \\ \hline
3           & Non-trainable Parameter         & 42,464                             \\ \hline
4           & Batch Size                       & 16                                 \\ \hline
5           & Epoch                            & 100                                \\ \hline
6           & Loss function                    & Sparse\_categorical\_crossentropy  \\ \hline
7           & Optimizer                        & Adam                               \\ \hline
8           & Activation Function              & SoftMax                            \\ \hline
9           & Learning Rate                    & 0.001                              \\ \hline
\end{tabular}
\end{table}
\subsection{Ablation Study}
The proposed EfficientNetB0ConvLSTM model demonstrates superior performance, achieving the highest accuracy of 89.22\%, with a precision of 88.12\%, recall of 86.54\%, and an F1 score of 87.96\% across the selected six classes. In comparison, other well-known models were evaluated to benchmark their effectiveness. The LSTM model achieved an accuracy of 76.44\%, with a precision of 75.96\%, a recall of 76.42\%, and an F1 score of 78.18\%. The ConvLSTM model performed with an accuracy of 70.35\%, precision of 71.07\%, recall of 73.33\%, and an F1 score of 69.28\%. Similarly, the EfficientNetB0 model attained an accuracy of 72.57\%, precision of 74.24\%, recall of 72.57\%, and an F1 score of 75.45\%. The broader EfficientNet (B0-B7) model achieved an accuracy of 76.75\%, with precision of 74.48\%, recall of 71.73\%, and an F1 score of 73.65\%. These results, summarized in Table \ref{table:ablation_study}, underscore the efficacy of the proposed EfficientNetB0ConvLSTM model in outperforming other baseline methods for the HMDB51 dataset.

\begin{table}[]
\caption{Ablation study of the proposed model with HMDB15 dataset} \label{table:ablation_study}
\centering
\begin{tabular}{|p{2cm}|l|l|l|l|}
\hline
\textbf{Methods} & \textbf{Accuracy} & \textbf{Precision} & \textbf{Recall} & \textbf{F1-Score} \\ \hline
LSTM & 76.44 & 75.96 & 76.42 & 78.18 \\ \hline
ConvLSTM & 70.35 & 71.07 & 73.33 & 69.28 \\ \hline
EfficientNetB0 & 72.57 & 74.24 & 72.57 & 75.45 \\ \hline
EfficientNet (B0-B7) & 76.75 & 74.48 & 71.73 & 73.65 \\ \hline
Proposed (EfficientNetB + ConvLSTM) & 89.22 & 88.12 & 86.54 & 87.96 \\ \hline
\end{tabular}
\label{tab:ablation_study}
\end{table}

\subsection{Model Parameters List}
\begin{figure*}[ht]
\centering
    \begin{subfigure}[b]{0.36\textwidth}
        \centering
        \includegraphics[width=\textwidth]{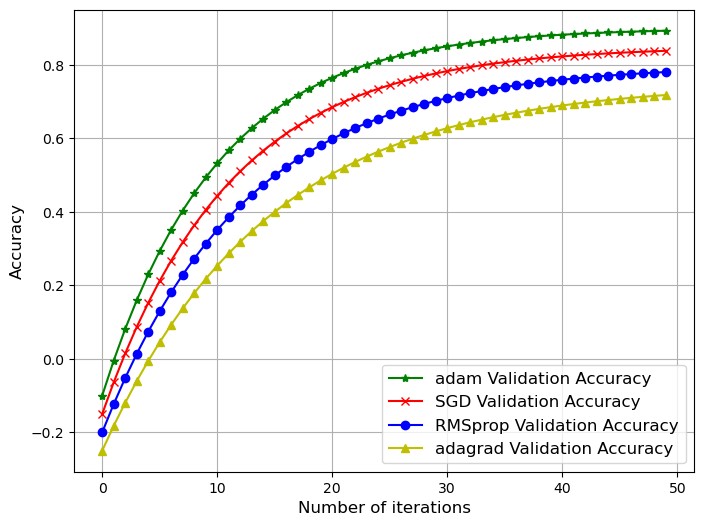}
        \caption{Accuracy curve}
      \end{subfigure}
    \begin{subfigure}[b]{0.36\textwidth}
        \centering
        \includegraphics[width=\textwidth]{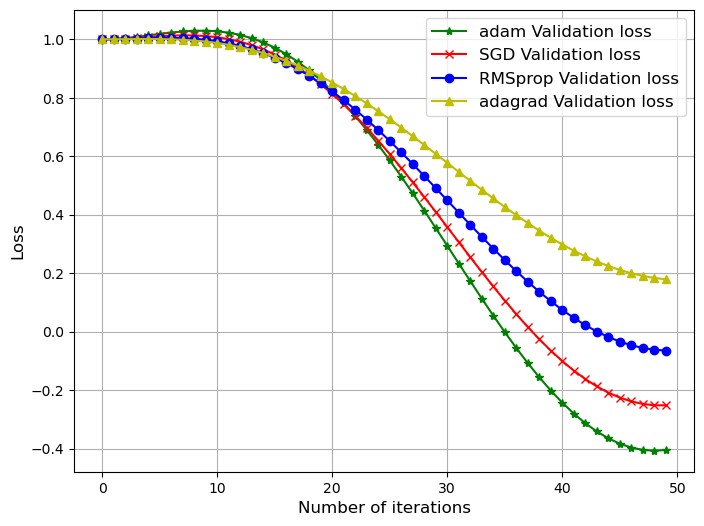}
        \caption{Loss curve}
      \end{subfigure}
    \captionsetup{justification=centering}
    \caption{Model optimizers comparison (a) Validation Accuracy and (b) Validation loss curve for NTU dataset.}
    \label{fig:ntu_accuracy_loss_curve}
\end{figure*}

\begin{figure*}[ht]
\centering
    \begin{subfigure}[b]{0.36\textwidth}
        \centering
        \includegraphics[width=\textwidth]{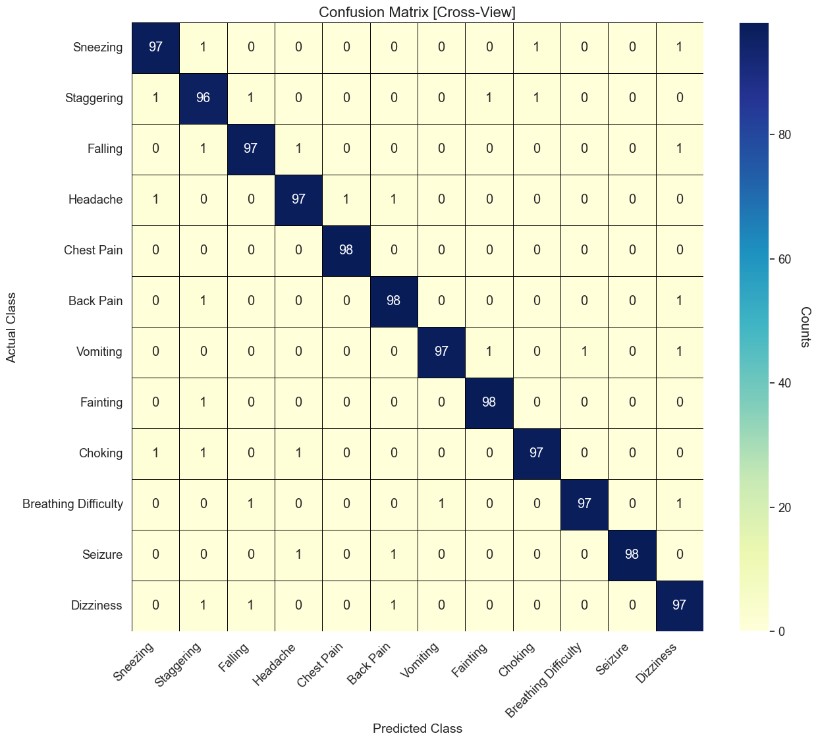}
        \caption{Confusion matrix (Cross-View) }
      \end{subfigure}
    \begin{subfigure}[b]{0.36\textwidth}
        \centering
        \includegraphics[width=\textwidth]{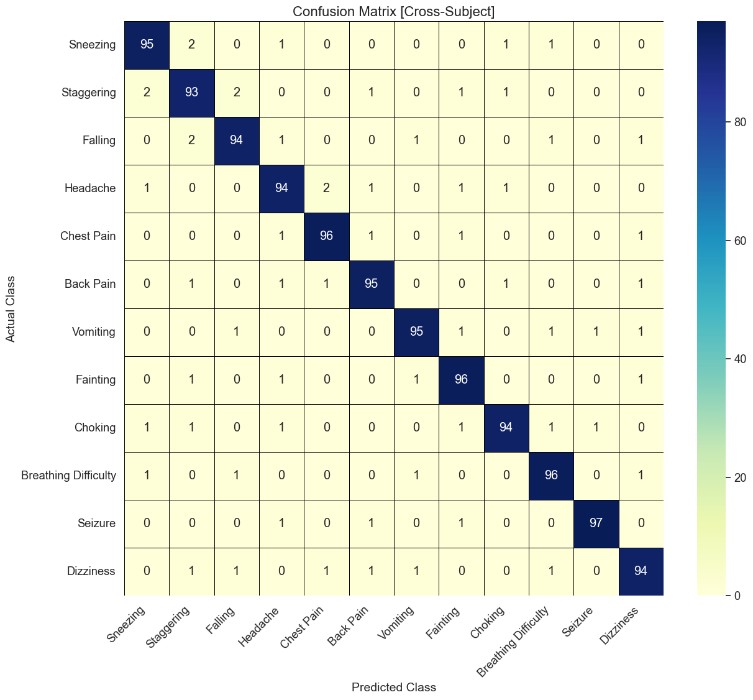}
        \caption{Model Loss (Cross-Subject)}
      \end{subfigure}
    \captionsetup{justification=centering}
    \caption{Confusion matrix (Cross-View) and (b) Model Loss (Cross-Subject) for NTU dataset.}
    \label{fig:ntu_confusion_matrix}
\end{figure*}

The proposed model has 8,138,104 parameters, of which 8,095,640 are trainable and 42,464 are non-trainable. The model is trained with a batch size of 16 for 100 epochs using sparse categorical cross-entropy as the loss function and Adam as the optimizer, with a learning rate of 0.001. The activation function is SoftMax, and Table \ref{tab:par_enconvlstm} lists the model's parameters. Figure \ref{fig:ntu_accuracy_loss_curve} highlights Adam's superior performance over SGD, RMSprop, and Adagrad in training the proposed model. Adam achieves the highest validation accuracy (approaching 0.95) and the lowest validation loss due to its adaptive learning rate mechanism, which accelerates convergence and improves generalization. The smooth accuracy and loss curves further demonstrate Adam’s stability and effectiveness, whereas other optimizers show slower convergence and higher validation losses. These results confirm Adam as the most effective optimizer for this model.
Figures \ref{fig:ntu_confusion_matrix} comprehensively evaluate the model's performance. The confusion matrix (Figure \ref{fig:ntu_confusion_matrix}) demonstrates robust classification accuracy. 

\subsection{Performance matrix with NTU RGB+D 120 Dataset}
The overall accuracy of this dataset is listed in Table \ref{tab:NTU_sota_accuracy}. And its Result in Comparison with Other Used Models. This comparison highlights the strength of your ENConvLSTM model in Human Motion Recognition (HMR) tasks, particularly in improving overall accuracy across both challenging metrics.

\subsubsection{Cross-Subject Evaluation} Involves splitting the dataset such that training and testing samples are from different subjects, ensuring the model generalizes well to unseen individuals. The model achieved a cross-subject accuracy of 94.85\%.

\subsubsection{Cross-View Evaluation} Involves splitting the dataset based on different camera views, ensuring the model performs well across various perspectives. The model achieved a cross-view accuracy of 96.45\%.
Key performance metrics—accuracy, precision, recall, and F1-score—are calculated for both cross-subject and cross-view evaluations, as shown in Table \ref{tab:NTU_crosssub_crosv_accuracy} .

\begin{table}[]
\caption{Result in Comparison with Other Used Models (For Overall Accuracy)} \label{tab:NTU_sota_accuracy}

\begin{tabular}{|l|l|l|}
\hline
Previous   Used Methods & Cross-Subject   (\%) & Cross-View   (\%) \\ \hline
Dynamic   Skeleton {[}9{]} & 60.2 & 65.2 \\ \hline
P-LSTM   {[}10{]} & 62.9 & 70.3 \\ \hline
STA-LSTM   {[}11{]} & 73.4 & 81.2 \\ \hline
TCN {[}12{]} & 74.3 & 83.1 \\ \hline
VA-LSTM   {[}13{]} & 79.2 & 87.7 \\ \hline
Deep   STGCK {[}14{]} & 74.9 & 86.3 \\ \hline
ST-GCN   {[}15{]} & 81.5 & 88.3 \\\hline
DPRL   {[}16{]} & 83.5 & 89.8 \\\hline
SR-TSL   {[}17{]} & 84.8 & 92.4 \\\hline
STGR-GCN   {[}18{]} & 86.9 & 92.3 \\\hline
AS-GCN   {[}19{]} & 86.8 & 94.2 \\\hline
2S-AGCN   {[}20{]} & 88.5 & 95.1 \\\hline
2sAGC-LSTM   {[}21{]} & 89.2 & 95.0 \\\hline
NAS-GCN   {[}22{]} & 89.4 & 95.7 \\\hline
ST-GCN   {[}23{]} & 90.1 & 95.9 \\\hline
Ours   (ENConvLSTM) & 94.85 & 96.45 \\ \hline
\end{tabular}
\end{table}

\begin{table}[]
\caption{Other performance evaluation metrics on “NTU RGB+D 120” Dataset} \label{tab:NTU_crosssub_crosv_accuracy}
\begin{tabular}{|l|l|l|l|}
\hline
\multirow{2}{*}{SN} & \multirow{2}{*}{\begin{tabular}[c]{@{}l@{}}Performance Martics\end{tabular}} & \multicolumn{2}{l|}{Dataset evaluation methods} \\ 
 &  & Cross-Subject (\%) & Cross-View (\%) \\ \hline
1 & Accuracy & 94.85\% & 96.45\% \\ \hline
2 & Precision & 93.70\% & 95.90\% \\ \hline
3 & Recall & 94.30\% & 96.10\% \\ \hline
4 & F1-Score & 94.00\% & 96.00\% \\ \hline
\end{tabular}
\end{table}

Below is a summary of Table \ref{tab:NTU_accuracy_loss} the accuracy and loss for both cross-view and cross-subject evaluations, followed by the corresponding hypothetical accuracy and loss curves.

\begin{table}[h!]
\centering
\caption{Accuracy and Loss Summary} 
\label{tab:NTU_accuracy_loss}
\begin{tabular}{|p{2cm}|p{3cm}|p{3cm}|}
\hline
\textbf{Metric}           & \textbf{Cross-Subject}             & \textbf{Cross-View}                \\ \hline
\textbf{Accuracy Summary} & Initial Accuracy: 75.01\%          & Initial Accuracy: 78.34\%          \\ 
                          & Final Accuracy: 94.85\%           & Final Accuracy: 96.45\%            \\ \hline
\textbf{Loss Summary}     & Initial Loss: 0.60                & Initial Loss: 0.55                \\ 
                          & Final Loss: 0.04                 & Final Loss: 0.03                 \\ \hline
\end{tabular}
\end{table}

In the cross-subject evaluation, the accuracy exhibits a robust upward trend from an initial 75.01\% to a final 96.85\%, with some minor fluctuations and eventual stabilization. Concurrently, the loss decreases consistently from 0.60 to 0.04, reflecting an overall improvement in model performance. Similarly, the cross-view evaluation shows a strong accuracy progression, starting at 78.34\% and reaching 96.45\% by the end, with minor variations throughout and a stabilizing trend. The loss trend in this evaluation mirrors that of the cross-subject assessment, declining from 0.55 to 0.03, indicating effective learning and convergence. Both evaluations demonstrate an effective model performance enhancement over time, with accuracy improving and loss decreasing, culminating in stable performance metrics.




\begin{figure}[ht]
\centering
\includegraphics[scale=0.20]{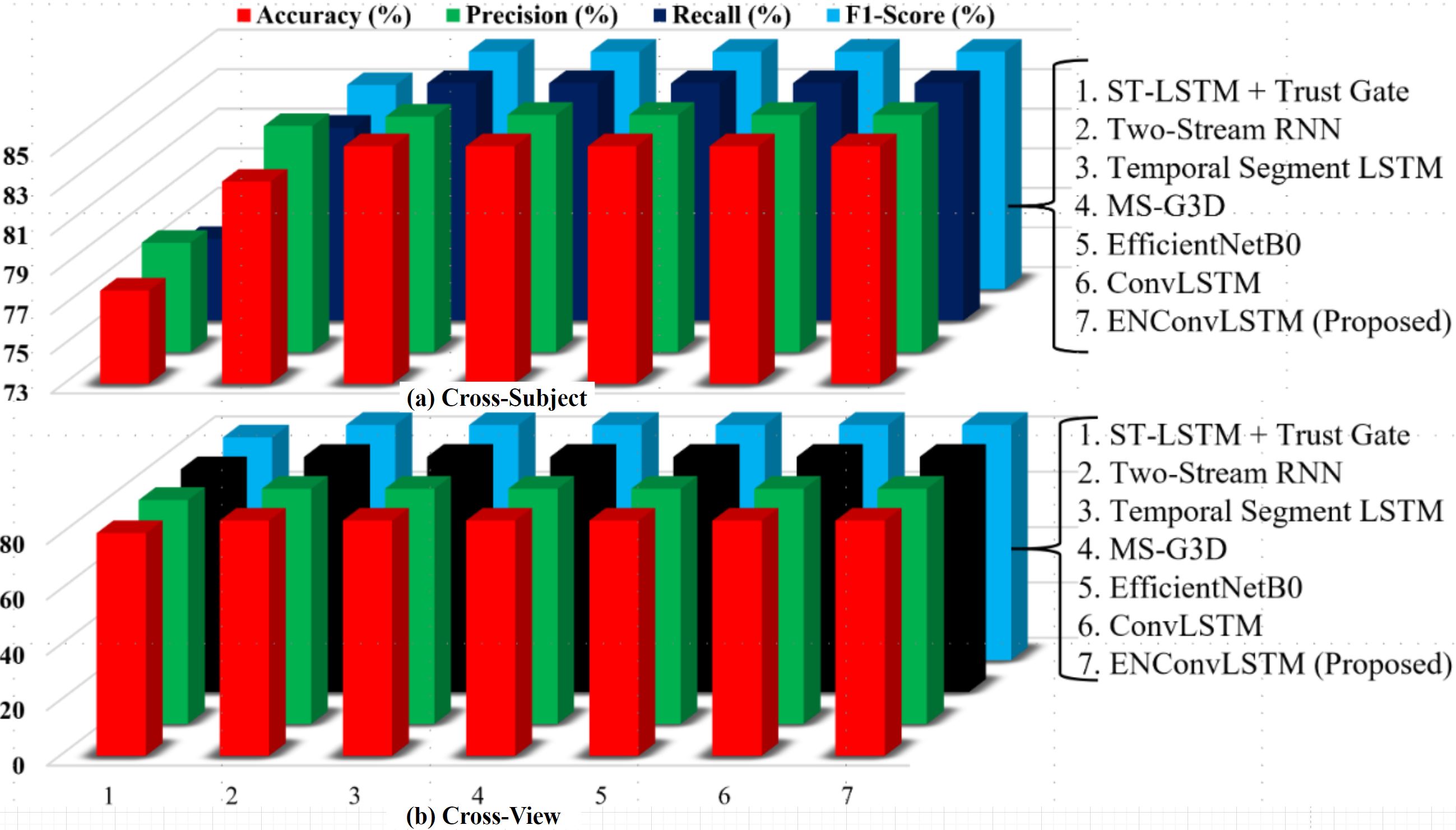}
\caption{Performance comparison among different tested models on the “NTU RGB+D 120” dataset with  (a) cross-view (b) cross-subject configuration. 
}
\label{fig:ntu_sota_cross_view}
\end{figure}



\begin{figure*}[ht]
\centering
\includegraphics[scale=0.28]{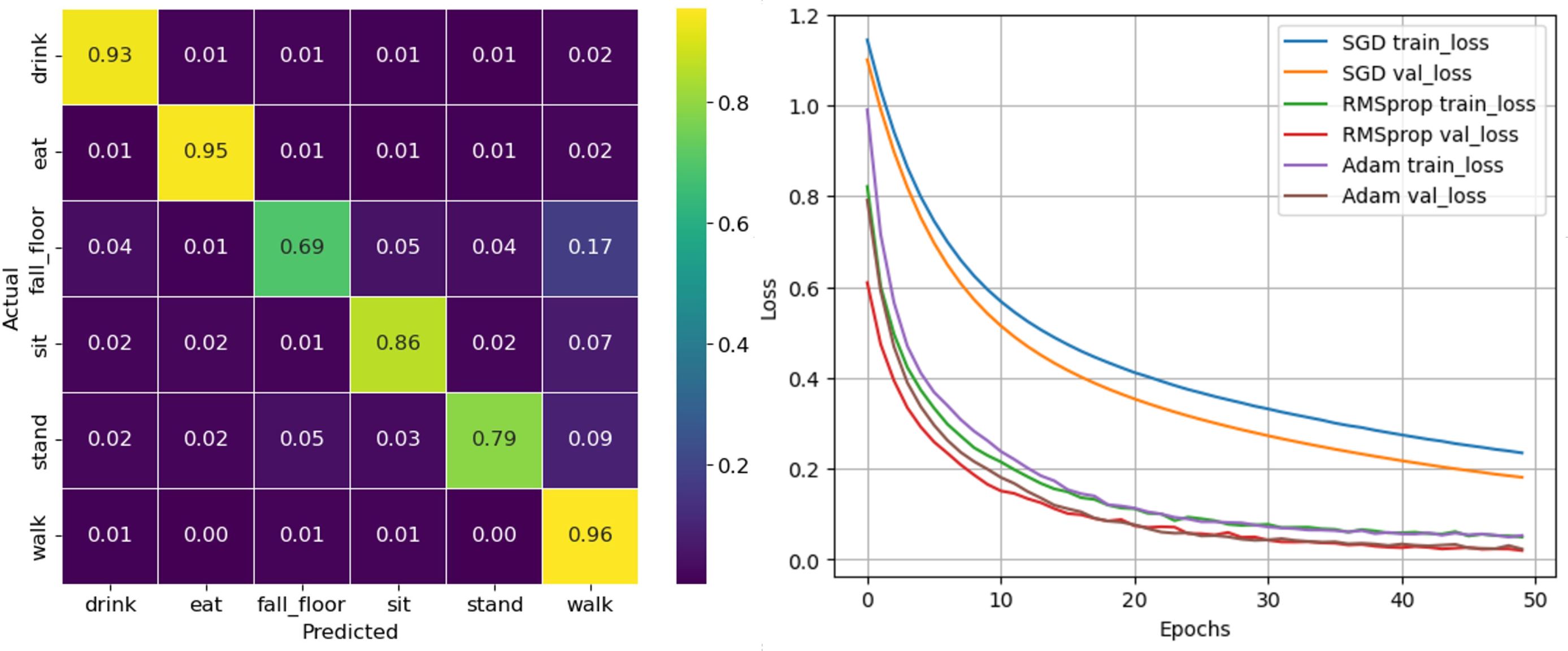}
\caption{ (a) Confusion matrix (b) Loss curve for the HMDB51 dataset. 
}
\label{fig:hmdb-conf-loss}
\end{figure*}

\begin{figure}[ht]
\centering
\includegraphics[scale=0.55]{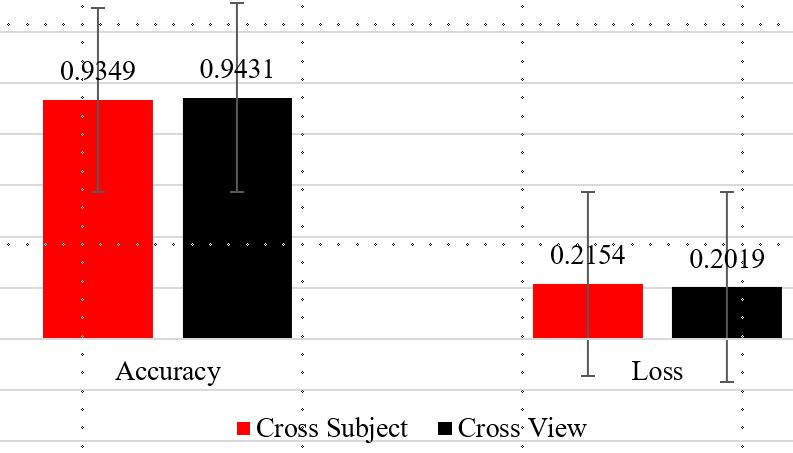}
\caption{Real-Time Human Motion Recognition Performance Metrics (Includes Accuracy and Loss Values). 
}
\label{fig:ntu_performance_real_time}
\end{figure}

\subsection{State of the art Comparison for the NTU RGB+D 120 Dataset}
The performance of the ENConvLSTM model is compared with several state-of-the-art models, including traditional methods and recent deep-learning approaches. The comparative analysis demonstrates the significant improvements achieved by the proposed model. The performance of the ENConvLSTM model is compared with several state-of-the-art models, including traditional methods and recent deep-learning approaches. The comparative analysis demonstrates in Table \ref{tab:NTU_sota_tow_ways} that significant improvements are achieved by the proposed model. Each Performance Evaluation Method (Cross-View and Cross-Subject) is graphically presented in Figure \ref{fig:ntu_sota_cross_view} (a) and Figure \ref{fig:ntu_sota_cross_view}(b). The proposed ENConvLSTM model significantly outperforms the other models across cross-subject and cross-view evaluations on the NTU RGB+D 120 dataset, showing excellent accuracy, precision, recall, and F1 score results. This indicates that the proposed model is highly robust and performs well across different subjects and viewing conditions.
\begin{table*}[]
\caption{“NTU RGB+D 120” dataset Accuracy summary list} \label{tab:NTU_sota_tow_ways}
\begin{tabular}{|l|l|l|l|l|l|l|}
\hline
Used Model with citation & \begin{tabular}[c]{@{}l@{}}Dataset\\    \\ Name\end{tabular} & Evaluation Method & Accuracy (\%) & Precision (\%) & Recall (\%) & F1-Score (\%) \\
ST-LSTM + Trust Gate & NTU RGB+D 120 & Cross-Subject & 77.71 & 78.54 & 77.14 & 77.81 \\\hline
Two-Stream RNN &NTU RGB+D 120 & Cross-Subject& 83.22 & 84.43 & 82.76 & 83.32 \\\hline
Temporal Segment LSTM &NTU RGB+D 120 & Cross-Subject& 85.23 & 84.91 & 85.18 & 85.77 \\\hline
MS-G3D & NTU RGB+D 120 & Cross-Subject& 89.44 & 89.15 & 89.33 & 89.23 \\ \hline
EfficientNetB0 &NTU RGB+D 120 & Cross-Subject & 92.55 & 91.86 & 92.45 & 92.15 \\\hline
ConvLSTM &NTU RGB+D 120 & Cross-Subject & 93.23 & 93.78 & 92.96 & 92.95 \\\hline
ENConvLSTM (Proposed) & NTU RGB+D 120 & Cross-Subject  & 94.85 & 93.72 & 94.31 & 94.76 \\\hline
ST-LSTM + Trust Gate &NTU RGB+D 120  &Cross-View & 80.47 & 80.91 & 80.54 & 80.48 \\
Two-Stream RNN &NTU RGB+D 120  &Cross-View & 85.24 & 85.54 & 84.95 & 85.25 \\\hline
Temporal Segment LSTM & NTU RGB+D 120  &Cross-View & 87.82 & 87.26 & 87.63 & 87.42 \\\hline
MS-G3D & NTU RGB+D 120  &Cross-View& 90.25 & 90.31 & 90.15 & 90.05 \\ \hline
EfficientNet &NTU RGB+D 120  &Cross-View& 93.17 & 92.73 & 93.54 & 92.85 \\\hline
ConvLSTM & NTU RGB+D 120  &Cross-View& 94.02 & 94.16 & 93.84 & 93.95 \\\hline
Proposed (ENConvLSTM) &NTU RGB+D 120  & Cross View & 96.45 & 95.91 & 96.17 & 96.32 \\ \hline
\end{tabular}
\end{table*}

\begin{figure}[ht]
\centering
\includegraphics[scale=0.28]{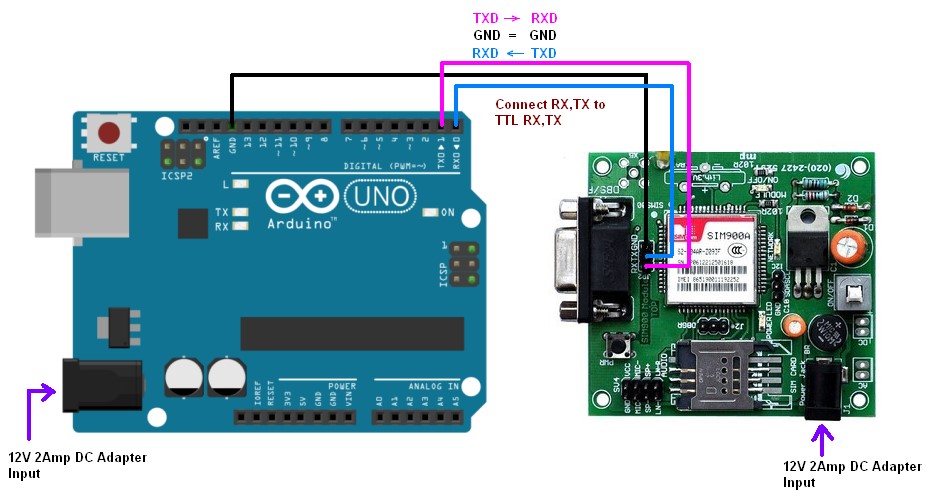}
\caption{Schematic diagram illustrating the connections between an Arduino UNO and a SIM900 GSM TTL module, highlighting the power supply, TX and RX pin connections, and the common ground to facilitate communication for tasks such as SMS sending and voice calling. 
}
\label{fig:uno_sim00}
\end{figure} 

\begin{figure}[ht]
\centering
\includegraphics[scale=0.20]{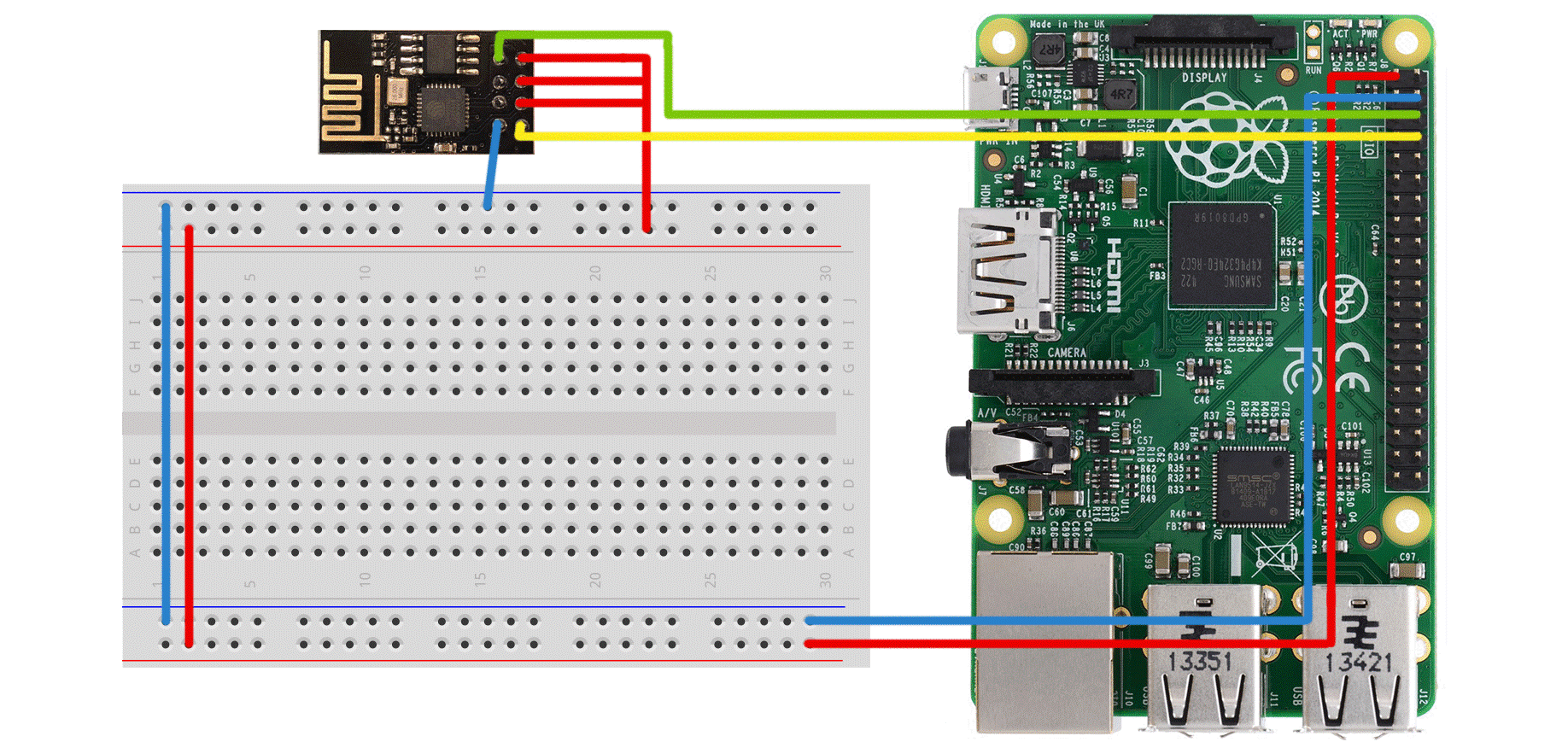}
\caption{Schematic diagram showing the wiring connections between a Raspberry Pi and an ESP8266 Wi-Fi module using a breadboard, illustrating power (VCC and GND) and data transmission (TX and RX) connections for enabling wireless communication in IoT projects. 
}
\label{fig:pi_wifi}
\end{figure} 
\subsection{Perfomrance Accuracy with HMDB51 Dataset}
Figure~\ref{fig:hmdb-conf-loss} presents the confusion matrix and loss with various optimizations. Therefore, we consider the Adam optimizer for this study.
This matrix offers a detailed view of accuracy for each individual class. Figure \ref{fig:hmdb-conf-loss} (a) shows the confusion matrix of the proposed model for the HMDB15 dataset. From this matrix, we can see that the diagonal line represents the true classes, while the off-diagonal data represents false detections. The classes "drink," "eat," and "walk" achieve remarkable accuracy in this dataset. However, the other classes show comparatively lower accuracy. We analyze the validation loss curve relative to the number of iterations, as illustrated in Figure~\ref{fig:hmdb-conf-loss} (b). The curve indicates that the ADAM optimizer outperforms both the SGD and RMSProp algorithms. Based on the analysis, Adam is likely the best-performing optimizer among the three optimizers, such as Adam~\cite{paul2024adam}, SGD~\cite{utomo2024hiragana,merrouchi2024autolropt}, and RMSProp~\cite{utomo2024hiragana,ahda2024comparison}. This is because Adam combines the advantages of both RMSProp and momentum, leading to faster and more reliable convergence~\cite{utomo2024hiragana}. It adapts the learning rate during training and uses past gradient information to accelerate learning, making it efficient and effective for many deep learning problems~\cite{ahda2024comparison}. 

These standard classification metrics for the HMDB51 dataset are listed in Table \ref{tab:hmdb_accuracy_table}, where the average accuracy in our experiment is 89.22\%.

\begin{table}[]
\caption{Classification result for the HMDB51 dataset} \label{tab:hmdb_accuracy_table}
\begin{tabular}{|l|l|l|l|l|}
\hline
Selected Class Levels & Precision & Recall & F1-Score & Accuracy \\ \hline
drink & 89.00 & 93.00 & 91.00 & - \\ \hline
eat & 89.00 & 95.00 & 92.00 & - \\ \hline
fall\_floor & 84.00 & 69.00 & 76.00 & - \\ \hline
sit & 85.00 & 86.00 & 86.00 & - \\ \hline
stand & 91.00 & 79.00 & 85.00 & - \\ \hline
walk & 91.00 & 96.00 & 93.00 & - \\ \hline
Average & 89.00 & 89.00 & 89.00 & 89.22 \\ \hline
\end{tabular}
\end{table}

Table \ref{tab:sota_hmdb1} provides a comparative analysis of various models tested on the HMDB51 dataset for human action recognition. The proposed model, a multi-stage model, achieves the highest accuracy of 89.22\% , showcasing its superior performance. Among other competitive models, Bidirectional LSTM attains an accuracy of 76.30\% \cite{hassan2024bidirectional}, and ViT+PBiLSTM+DSMHA achieves 78.62\% \cite{hussain2024shots}. Models such as STM \cite{jiang2024stm} and Attention-Based LSTM with 3D CNN \cite{saoudi2023advancing} demonstrate strong performance with accuracies of 80.40\% and 87.98\%, respectively, while EfficientNet delivers an impressive accuracy of 88.70\% \cite{burton2024activation}. In contrast, earlier models like VicTR (ViT-B/16) \cite{kahatapitiya2024victr} and SVT (Fine-tune) \cite{ranasinghe2022transformer} achieve relatively lower accuracies of 51.00\% and 67.28\%, respectively. Additionally, the SVT (Linear) \cite{ranasinghe2022transformer} and LSTM \cite{hussain2022vision} models exhibit accuracies of 57.8\% and 73.75\%, highlighting the advancements made by recent approaches over traditional methodologies.

\begin{table}[h]
    \centering
    \caption{State-of-the-art comparison for the proposed model with HMDB51 dataset.} 
    \label{tab:sota_hmdb1}
    \begin{tabular}{|p{2.5cm}|p{3.1cm}|p{.7cm}|p{.7cm}|}
        \hline
        \textbf{Author and Citation} & \textbf{Model Name} & \textbf{Year} & \textbf{Accuracy (\%)} \\ \hline
        Liu et al. \cite{liu2017hierarchical} & Hierarchical Clustering Multi-Task Learning & 2017 & 51.41 \\ \hline
        Wang et al. \cite{wang2018twostream} & Two-Stream 3DCNN & 2018 & 70.53 \\ \hline
        Ullah et al. \cite{ullah2019activity} & Temporal Optical Flow LSTM & 2019 & 72.25 \\ \hline
        Wang et al. \cite{wang2019temporal} & Temporal Segment Networks & 2019 & 70.77 \\ \hline
        Ullah et al. \cite{ullah2019action} & Deep Autoencoder CNN & 2019 & 70.37 \\ \hline
        Ma et al. \cite{ma2019tslstm} & TS-LSTM & 2019 & 74.86 \\ \hline
        Majd et al. \cite{majd2020correlational} & Correlational ConvLSTM & 2020 & 66.28 \\ \hline
        He et al. \cite{he2021dblstm} & DB-LSTM & 2021 & 75.17 \\ \hline
        Hussain et al. \cite{hussain2022vision} & ViT & 2021 & 73.75 \\ \hline
        Xiao et al. \cite{feng2023graph} & Temporal Gradient Learning & 2022 & 75.96 \\ \hline
        Ranasinghe et al. \cite{ranasinghe2022transformer} & Self-Supervised Transf & 2022 & 67.28, 57.8 \\ \hline
        Sarraf et al.  \cite{sarraf2023transformer} & ViT & 2023 & 59.74, 68.2 \\ \hline
        Ullah et al. \cite{ullah2023recognition} & 2-Attention CNN-GRU & 2023 & 79.38 \\ \hline
        Saoudi et al. \cite{saoudi2023advancing} & Attention-LSTM-3DCNN & 2023 & 87.98 \\ \hline
        Burton-Barr et al. \cite{burton2024activation} & Act-Control Vision & 2024 & 88.70 \\ \hline
        Kahatapitiya et al. \cite{kahatapitiya2024victr} & VicTR & 2024 & 51.00 \\ \hline
        Hassan et al. \cite{hassan2024bidirectional} & D Bi-LSTM Transfer Learning & 2024 & 76.30 \\ \hline
        Hussain et al. \cite{hussain2024shots} & Dual-Stream Framework & 2024 & 78.62 \\ \hline
        Jiang et al. \cite{jiang2024stm} & STM Framework & 2024 & 80.40 \\ \hline
        \textbf{Proposed} & EfficientNetB0ConvLSTM & 2024 & \textbf{89.22} \\ \hline
    \end{tabular}
\end{table}

\section{Real-Deployment}
Finally, the proposed deep learning system is integrated with a Raspberry Pi and GSM module to send real-time alerts via Twilio’s SMS service, keeping caregivers and patients informed instantly. This system is scalable, efficient, and proactive, helping to improve patient monitoring and outcomes and reduce healthcare costs. 
\subsection{Real Time Implementation}
Figure \ref{fig:uno_sim00} shows the interconnection between an Arduino UNO and the SIM900 GSM module to implement the real-time scenario. A 12V 2Amp DC adapter powers the Arduino. The TX (transmit) and RX (receive) pins are crucial for communication. The TX pin of the Arduino connects to the RXD pin of the TTL module for data transmission, while the RX pin of the Arduino connects to the RDX pin of the TTL module for receiving data. Both devices share a common ground (GND) for proper operation. This setup enables the Arduino to communicate with the GSM module for tasks like sending SMS or making voice calls.

Figure \ref{fig:pi_wifi} illustrates the wiring connections between a Raspberry Pi and an ESP8266 Wi-Fi module using a breadboard. The connections include power and ground lines: the red wire represents the VCC (power) connection from the Raspberry Pi to the ESP8266, while the blue wire indicates the ground (GND) connection. The yellow wire connects the GPIO pin from the Raspberry Pi to the ESP8266's TX pin for data transmission, and the green wire connects another GPIO pin to the RX pin of the ESP8266 for data reception. This setup enables the Raspberry Pi to communicate with the ESP8266 for wireless connectivity in various projects, such as IoT applications. Arduino UNO REV3 - Compact and Versatile Microcontroller Board with A000066. Interface a SIM900A GSM module with an Arduino to send and receive SMS. Arduino UNO REV3 - Compact and Versatile Microcontroller Board with A000066. Interface a SIM900A GSM module with an Arduino to send and receive SMS.
Figure \ref{fig:proposed_dl_diagram} illustrates the proposed system for human motion recognition using an ENConvLSTM model integrated with an Arduino. It begins with an input video that is processed to extract skeleton data, represented in a 3D coordinate system (X, Y, Z). This skeleton data is fed into the ENConvLSTM model, which then utilizes a SoftMax layer to predict the activity being performed. The predicted activity is communicated to an Arduino, which can be powered by a stable 5V source or battery, indicating a real-time or recorded data application. This setup enables effective monitoring and recognition of human activities through skeletal motion analysis. The proposed human motion recognition system's performance in the laboratory experiment is satisfactory. However, there are always some differences between laboratory and real-time scenarios, table IX illustrates the result in real-time scenario, and their visual representation is shown in Figure \ref{fig:ntu_performance_real_time}.

\begin{table}[]
\caption{Labraory and Real-time experiment performance for the HMDB51 dataset} \label{tab:hmdb_result}
\begin{tabular}{|p{.2cm}|p{.7cm}|p{1.2cm}|p{1cm}|p{1cm}|p{.7cm}|p{.9cm}|}
\hline
No & Actual Class & Real-time observation (Individual) & Laboratory   Experiment (\%) & Real-time Experiment (\%) &Calculate score (\%) &Loss of  Real-time Experiment (\%)\\ \hline
1 & fall\_floor & Individual falling   to the floor & 89.00 & 87.00 & 97.75 & 2.00 \\ \hline
2 & walk & Individual walking & 95.00 & 91.00 & 95.79 & 4.00 \\ \hline
3 & stand & Individual standing & 79.00 & 77.00 & 97.47 & 2.00 \\ \hline
4 & eat & individual eating & 94.00 & 90.00 & 95.74 & 4.00 \\ \hline
5 & sit & individual sitting & 86.00 & 82.00 & 95.35 & 4.00 \\ \hline
6 & drink & individual drinking & 92.00 & 88.00 & 95.65 & 4.00 \\ \hline
\multicolumn{3}{l}{Average Score} & 89.22 & 85.83 & 96.29 & 3.33 \\ \hline
\end{tabular}
\end{table}

\begin{figure}[ht]
\centering
\includegraphics[scale=1.40]{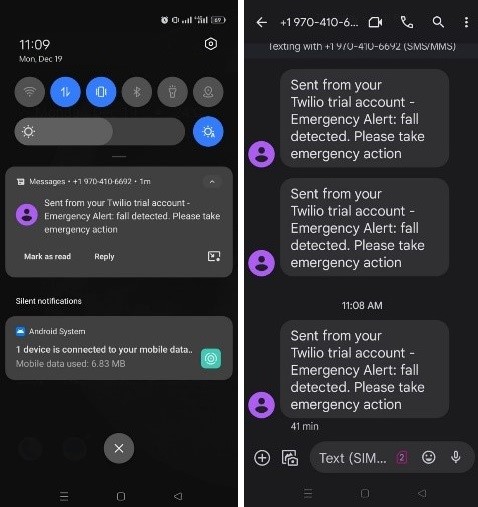}
\caption{Snapshots of warning messages. 
}
\label{fig:output_message}
\end{figure}

\subsection{Sending Alert Message (Notification)}
In our proposed system, the model predicts human actions, specifically identifying whether an individual is experiencing a fall. Upon detecting a fall event, the system promptly triggers an alert message to the registered user’s mobile device. Figure \ref{fig:output_message} presents snapshots of the warning messages generated by the system upon detecting a fall event. This real-time communication is facilitated through the integration of Twilio, a leading SMS and communication server provider, ensuring reliable and instantaneous notifications. By leveraging Twilio's robust platform, we enhance the responsiveness of our human motion recognition system, thereby significantly improving user safety and emergency response efficiency.

\section{Conclusion and Future Work}
\label {sec: Conclusion and future plan}This study introduces a novel IoT-based framework for real-time Medical-Related Human Activity (MRHA) recognition, addressing critical challenges such as high computational demands, low accuracy, and adaptability limitations in existing systems. By combining EfficientNet for spatial feature extraction with ConvLSTM for spatio-temporal feature integration, the proposed method achieves robust performance, effectively recognizing key MRHA activities. The system demonstrated high accuracy on the NTU RGB+D 120 dataset, achieving 94.85\% for cross-subject and 96.45\% for cross-view evaluations, as well as 89.00\% accuracy on the HMDB51 dataset. These results validate the model's efficiency in recognizing healthcare-related activities such as sneezing, falling, walking, and sitting. A major contribution of this research is the integration of the MRHA recognition system with a Raspberry Pi and GSM module, enabling real-time alerts via Twilio’s SMS service. This design eliminates reliance on third-party applications, ensuring instant notifications for caregivers and enhancing usability. The scalable system demonstrates significant potential for improving patient monitoring, healthcare outcomes, and cost-efficiency by providing proactive and responsive solutions. Despite its promising performance, the system may face challenges in real-time applications due to sensitivity to subject distance, varying environmental conditions, and background clutter. These factors can impact adaptability in dynamic real-world scenarios, including low-light and night-vision settings. To address these limitations, future work will focus on integrating multimodal datasets, incorporating RGB, Depth, and other sensor data to enhance robustness. Furthermore, leveraging cloud computing for large-scale remote monitoring and addressing privacy and security concerns for sensitive patient data are crucial for practical deployment. Additionally, incorporating AI-driven anomaly detection for real-time health alerts and collaborating with healthcare professionals will ensure the system's practicality and effectiveness in clinical environments. This research sets the foundation for future innovations in MRHA recognition, advancing the role of IoT and deep learning in real-time healthcare applications.
.


\section*{Acknowledgment}
I extend my sincere thanks to the Information and Communication Technology Division of the Ministry of Posts, Telecommunication, and Information Technology of Bangladesh, People's Republic of Bangladesh, for their invaluable support and funding of my ICT fellowship program in the MPhil Degree. Additionally, I would like to express my gratitude to my supervisor and co-authors for their guidance and contributions to this research.



\bibliographystyle{IEEEtran}
\bibliography{Mybib}


\begin{IEEEbiography}    [{\includegraphics[width=1in,height=1.25in,clip,keepaspectratio]{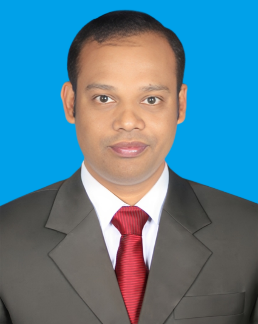}}]
{Subrata Kumer Paul} completed his B.Sc. and M.Sc. Engineering from Rajshahi University, in Computer Science and Engineering in 2016 and 2017, respectively. Now, he is working as an Assistant Professor at the Bangladesh Army University of Engineering \& Technology (BAUET), Qadirabad Cantonment, Natore-6431, Bangladesh. Also, he has been pursuing the MPhil Program at Rajshahi University. His research fields are Deep Learning, Data Mining, and Speech Signal Processing. He has published more than 25 international journal/book chapter/conference papers. Mr. Subrata received a fellowship from the Information and Communication Technology Division of the Ministry of Posts, Telecommunication, and Information Technology of Bangladesh. Mr. Subrata is also a Graduate member of IEEE.
\end{IEEEbiography}

\begin{IEEEbiography}    [{\includegraphics[width=1in,height=1.25in,clip,keepaspectratio]{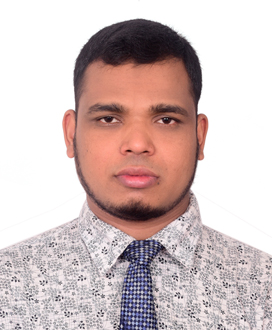}}]
{Abu Saleh Musa Miah} received the B.Sc.Engg. and M.Sc.Engg. degrees in computer science and engineering from the Department of Computer Science and Engineering, University of Rajshahi, Rajshahi-6205, Bangladesh, in 2014 and 2015, respectively, achieving the first merit position. He received his Ph.D. in computer science and engineering from the University of Aizu, Japan, in 2024, under a scholarship from the Japanese government (MEXT). He assumed the positions of Lecturer and Assistant Professor at the Department of Computer Science and Engineering, Bangladesh Army University of Science and Technology (BAUST), Saidpur, Bangladesh, in 2018 and 2021, respectively. Currently, he is working as a visiting researcher (postdoc) at the University of Aizu since April 1, 2024. His research interests include AI, ML, DL, Human Activity Recognition (HCR), Hand Gesture Recognition (HGR), Movement Disorder Detection, Parkinson's Disease (PD), HCI, BCI, and Neurological Disorder Detection. He has authored and co-authored more than 50 publications in widely cited journals and conferences.
\end{IEEEbiography}

\begin{IEEEbiography}    [{\includegraphics[width=1in,height=1.25in,clip,keepaspectratio]{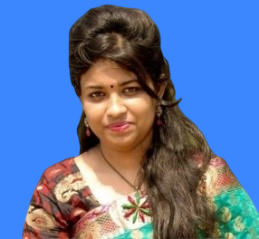}}]
{Rakhi Rani Paul} graduated with her B.Sc. and M.Sc. Engineering from Rajshahi University, in Computer Science and Engineering in 2017 and 2018, respectively. Now, she is working as a Lecturer at the Bangladesh Army University of Engineering \& Technology (BAUET), Qadirabad Cantonment, Natore-6431, Bangladesh. Her research fields are Deep Learning, Data Mining, and Speech Signal Processing.
\end{IEEEbiography}

\begin{IEEEbiography}[{\includegraphics[width=1in,height=1.25in, clip,keepaspectratio]{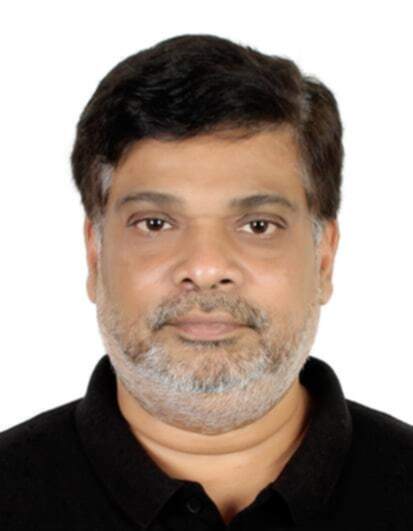}}]{
Md Ekramul Hamid} received the B.Sc. and M.Sc. degrees in Applied. 
Physics and Electronics from the University of Rajshahi. Later on, I received a Master in Computer Science degree from Pune University, India, and a PhD degree from Shizuoka University, Japan. He is currently working as a professor at the Department of Computer Science and Engineering, University of Rajshahi, Bangladesh. He has published more than 60 international journal/conference papers. He is a recipient of the Monbukagakusho scholarship, JASSO Fellowship, and NIST fellowship for his contribution to Science and Technology. He worked as a Faculty member at King Khalid University, KSA, in 2010- 11 and as a visiting researcher at Shizuoka University, Japan, in 2012, 2014, and 2017. He worked as the Chairman of the CSE department and the Dean of the Faculty of 
Engineering at the University of Rajshahi. His research interests include Digital signal processing, Machine learning, Analysis and synthesis of speech signals, Speech enhancement, and Image processing.
\end{IEEEbiography}

 \begin{IEEEbiography}[{\includegraphics[width=1in,height=1.25in, clip,keepaspectratio]{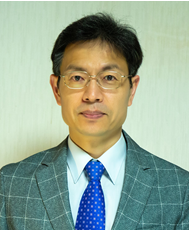}}] {
JUNGPIL SHIN} (Senior Member, IEEE) received the B.Sc. degree in computer science and statistics and the M.Sc. degree in computer science from Pusan National University, South Korea, in 1990 and 1994, respectively, and the Ph.D. degree in computer science and communication engineering from Kyushu University, Japan, in 1999, under a scholarship from the Japanese Government (MEXT). He was an Associate Professor, a Senior Associate Professor, and a Full Professor with the School of Computer Science and Engineering, The University of Aizu, Japan, in 1999, 2004, and 2019, respectively. He has co-authored more than 420 published papers for widely cited journals and conferences. His research interests include pattern recognition, image processing, computer vision, machine learning, human–computer interaction, non-touch interfaces, human gesture recognition, automatic control, Parkinson’s disease diagnosis, ADHD diagnosis, user authentication, machine intelligence, bioinformatics, and handwriting analysis, recognition, and synthesis. He is a member of ACM, IEICE, IPSJ, KISS, and KIPS. He serves as an Editorial Board Member for Scientific Reports. He was included among the top 2\% of scientists worldwide edition of Stanford University/Elsevier, in 2024. He served as the general chair, the program chair, and a committee member for numerous international conferences. He serves as an Editor for IEEE journals, Springer, Sage, Taylor \& Francis, Sensors (MDPI), Electronics (MDPI), and Tech Science. He serves as a reviewer for several major IEEE and SCI journals.
\end{IEEEbiography}

\begin{IEEEbiography}[{\includegraphics[width=1in,height=1.25in, clip,keepaspectratio]{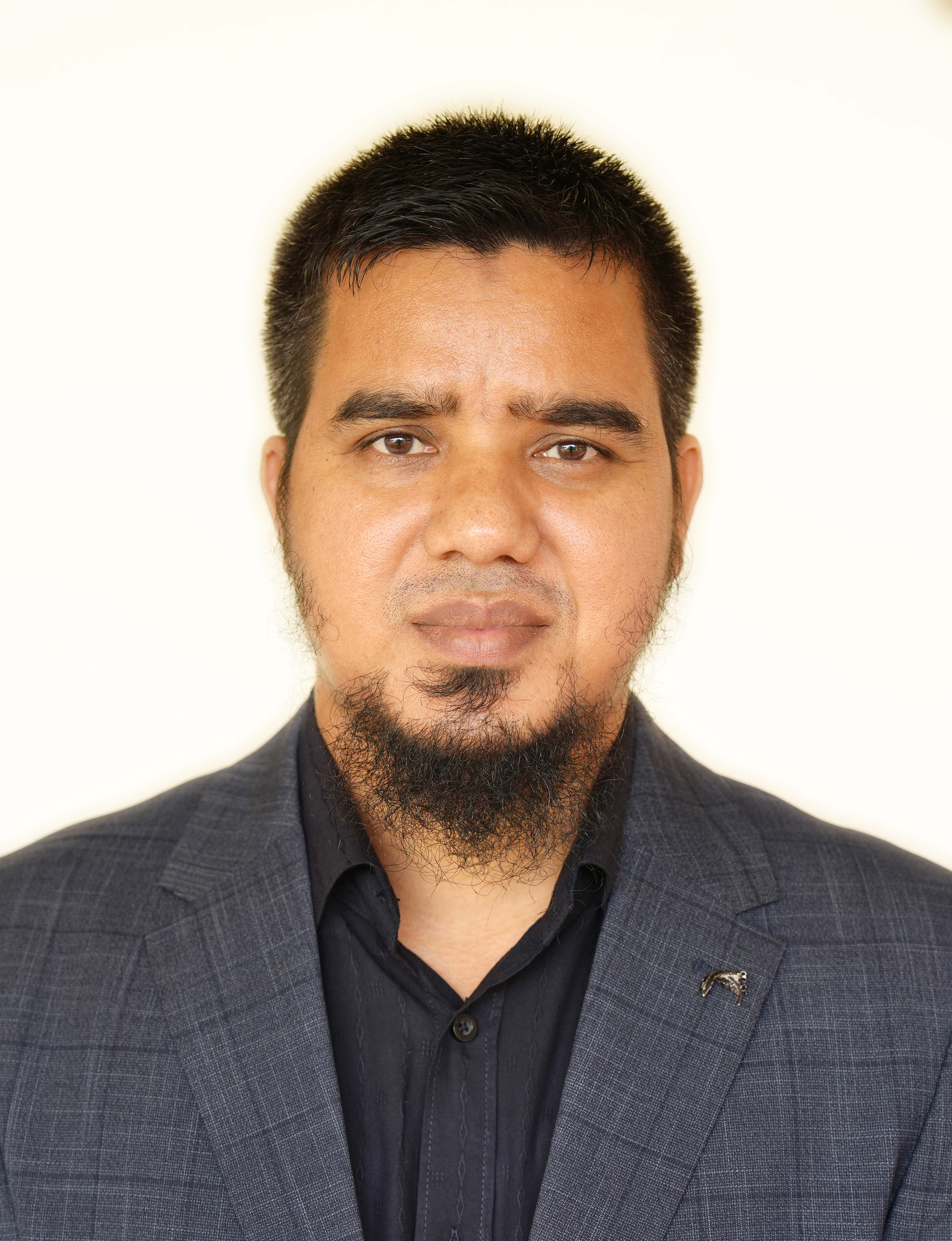}}] {
Md Abdur Rahim} received his Ph.D. in 2020 from the Graduate School of Computer Science and Engineering at The University of Aizu in Fukushima, Japan. He completed his Bachelor of Science (Honours) and Master of Science (M.Sc.) degrees in Computer Science and Engineering at the University of Rajshahi in Bangladesh, graduating in 2008 and 2009, respectively. Currently, he is an Associate Professor and the Head of the Department of Computer Science and Engineering at Pabna University of Science and Technology in Pabna 6600, Bangladesh. His research interests encompass human-computer interaction, pattern recognition, computer vision and image processing, human recognition, and machine intelligence. He has published numerous papers in major journals (SCI and SCIE indexed) and conferences, as well as reviews for several SCI/SCIE indexed journals and international conferences.
\end{IEEEbiography}

\EOD

\end{document}